\documentclass{article} % For LaTeX2e
\usepackage{iclr2026_conference,times}

\usepackage{amsmath,amsfonts,bm}

\def\eqref#1{equation~\ref{#1}}
\def\1{\bm{1}}

\DeclareMathAlphabet{\mathsfit}{\encodingdefault}{\sfdefault}{m}{sl}
\SetMathAlphabet{\mathsfit}{bold}{\encodingdefault}{\sfdefault}{bx}{n}

\usepackage{hyperref}
\usepackage{url}
\usepackage{graphicx}
\usepackage{subcaption}
\usepackage[section]{placeins} % keep floats within their section

\title{Bypassing the Rationale: Causal Auditing of Implicit Reasoning in Language Models}

\author{Anish Sathyanarayanan\textsuperscript{*}, Aditya Nagarsekar\textsuperscript{*}, Aarush Rathore\textsuperscript{*} \\
BITS Pilani, K. K. Birla Goa Campus \\
\texttt{\{f20240559, f20230473, f20230598\}@goa.bits-pilani.ac.in}
}

\fancypagestyle{equalcontribution}{%
  \fancyfoot[L]{\footnotesize\textsuperscript{*}Equal contribution.}%
}

\iclrfinalcopy % Camera-ready version

\begin{document}

\maketitle
\thispagestyle{equalcontribution}

\begin{abstract}
Chain-of-thought (CoT) prompting is widely used as a reasoning aid and is often treated as a transparency mechanism. Yet behavioral gains under CoT do not imply that the model’s internal computation causally depends on the emitted reasoning text, i.e. models may produce fluent rationales while routing decision-critical computation through latent pathways. We introduce a causal, layerwise audit of CoT faithfulness based on activation patching. Our key metric, the \emph{CoT Mediation Index} (CMI), isolates CoT-specific causal influence by comparing performance degradation from patching CoT-token hidden states against matched control patches. Across multiple model families (Phi, Qwen, DialoGPT) and scales, we find that CoT-specific influence is typically \emph{depth-localized} into narrow ``reasoning windows,'' and we identify \emph{bypass regimes} where CMI is near-zero despite plausible CoT text. We further observe that models tuned explicitly for reasoning tend to exhibit stronger and more structured mediation than larger untuned counterparts, while Mixture-of-Experts models show more distributed mediation consistent with routing-based computation. Overall, our results show that CoT faithfulness varies substantially across models and tasks and cannot be inferred from behavior alone, motivating causal, layerwise audits when using CoT as a transparency signal.
\end{abstract}

\section{Introduction}
\label{sec:intro}

Chain-of-thought (CoT) prompting has emerged as a powerful technique for improving the reasoning performance of large language models (LLMs) and for making their decisions more transparent. By encouraging models to produce intermediate reasoning steps before a final answer, CoT is often treated as a window into the model’s internal computation. In safety and alignment contexts, such reasoning traces are frequently interpreted as evidence that the model is following a faithful, interpretable decision process. However, an important mechanistic question remains unresolved: \emph{do models actually use chain-of-thought internally?} Behavioral evaluations alone cannot answer this. A model may produce coherent, task-relevant CoT while internally routing its computation through pathways that do not causally depend on the emitted reasoning tokens. In such cases, the visible CoT may function as a post-hoc rationalization or stylistic artifact rather than a faithful representation of the model’s internal reasoning.

To address this gap, we introduce a \textbf{causal, layerwise auditing framework} for evaluating CoT faithfulness. Instead of relying solely on output-level behavior, we intervene directly on internal activations and measure how these perturbations affect the model’s predictions. This lets us test which internal states matter for CoT Concretely, we use layerwise activation patching to measure how much CoT-related representations matter for the final answer. We summarize this with \emph{CoT Mediation Index} (CMI) across layers. Our findings reveal that CoT faithfulness varies substantially across models. In many cases, CMI is highly localized in depth, with only a small subset of layers carrying most of the CoT-specific causal effect. We also observe \emph{bypass regimes} in which models behaviorally emit plausible CoT while showing little additional mechanistic reliance on it, suggesting that visible reasoning traces can diverge from internal computation. Because transformers can move information across positions via attention, layerwise patching at CoT token positions can miss cases where the model reads CoT at earlier layers and writes the relevant information into non-CoT positions. In such cases, patching CoT positions at later layers can yield $\mathrm{CMI}_\ell \approx 0$ even though CoT causally influenced the computation earlier. Thus, low $\mathrm{CMI}_\ell$ should be interpreted as \emph{no remaining CoT-position mediation at layer $\ell$} under our intervention, which may point to complete CoT bypass, but not necessarily so.

\paragraph{Contributions.}
\begin{enumerate}
  \item We introduce a causal, layerwise intervention and a CoT-specific metric (CMI) to test whether models mechanistically rely on CoT.
  \item We localize where CoT mediation occurs in depth and use these profiles to compare models and identify routing regimes, including possible bypass.
  \item We show that behavior can mislead about CoT faithfulness, motivating causal audits.
\end{enumerate}

\section{Background and Related Work}
\label{sec:related}

Our work lies at the intersection of three threads: (i) the faithfulness of chain-of-thought (CoT) reasoning, (ii) strategic or evaluation-aware model behavior, and (iii) causal intervention methods for probing internal computation.

CoT is often treated as a transparency mechanism, yet models can optimize outputs for evaluation rather than faithfully expose internal computation~\citep{greenblatt2024alignmentfaking,sharma2023sycophancy}. A growing body of work shows that reasoning traces may be unfaithful or post-hoc, with plausible explanations that do not causally drive the final answer~\citep{yee2024dissociation,chen2025reasoning}. These issues are compounded in oversight settings: LLM-as-judge studies demonstrate that unfaithful CoT can systematically mislead text-based evaluators, revealing that both explanations and judges are gameable~\citep{khalifa2026gaming}.

Related work documents that models can produce strategic or unreliable reasoning traces, including alignment faking and evaluation-aware behavior~\citep{greenblatt2024alignmentfaking,baker2025monitoringreasoning,wang2025whenthinkingllmslie}. In these cases, CoT may be shaped to satisfy monitoring rather than reflect the internal process used to produce the answer. This motivates distinguishing between surface-level signals in the text and the internal mechanisms that determine model behavior. Prior faithfulness evaluations often rely on behavioral perturbations or text-level analyses, which can reveal inconsistencies but do not localize where (or whether) reasoning traces are mechanistically integrated. Our approach addresses this gap by pairing interpretable surface scoring with causal intervention methods inspired by activation patching and causal tracing in language models~\citep{heimersheim2024activation,zhang2023bestpractices}. Rather than asking whether explanations merely correlate with outputs, we test whether CoT-aligned internal representations are \emph{causally involved} in producing the answer.

Within this landscape, our contribution is a causal, layerwise auditing framework that measures whether answers are \emph{mechanistically mediated} by CoT-aligned internal representations. We quantify this dependence using the CoT Mediation Index (CMI) and localize where in the network CoT-specific influence arises. By contrasting these causal signals with behavioral CoT indicators, we directly test the central distinction raised in recent work: whether CoT serves as a post-hoc narrative or is functionally integrated into the model’s reasoning process.

\section{Method: Causal Intervention Framework}
\label{sec:method}

We treat CoT faithfulness as a mechanistic question: does the model's final answer actually route through the hidden states aligned with the CoT tokens? To quantify this, we use \emph{source patching} to estimate how much answer likelihood depends on CoT activations.

\subsection{Causal intervention setup}
\label{sec:method:cmi}

To isolate the causal effect of explicit reasoning, we compare a \emph{With-CoT} run $x_c$ against a \emph{No-CoT} run $x_{\neg c}$. We patch hidden states from $x_{\neg c}$ into $x_c$ at the CoT token positions and measure the resulting (non-negative) decrease in the log-probability of a reference answer. We report results per layer.
We define two primary metrics of sensitivity at layer $\ell$:
\begin{itemize}
\item \textbf{CoT Drop} ($\Delta_{\mathrm{cot},\ell}$): the (non-negative) decrease in answer log-probability when patching the hidden states at the CoT token positions in $x_c$ using activations from $x_{\neg c}$:
\begin{equation}
\Delta_{\mathrm{cot},\ell}
= \max\left(0,\ \log P(y \mid x_c) - \log P\bigl(y \mid \text{patch}_{\mathcal{C}}(x_c, x_{\neg c})\bigr)\right).
\end{equation}
\item \textbf{Control Drop} ($\Delta_{\mathrm{ctrl},\ell}$): a matched ``placebo'' intervention that patches a same-size set of random \emph{non-CoT} token positions:
\begin{equation}
\Delta_{\mathrm{ctrl},\ell}
= \max\left(0,\ \log P(y \mid x_c) - \log P\bigl(y \mid \text{patch}_{\mathcal{N}}(x_c, x_{\neg c})\bigr)\right).
\end{equation}
In practice, we average this quantity over multiple random draws to reduce variance.
\end{itemize}

\subsection{CoT Mediation Index (CMI) and Bypass}
\label{sec:method:cmi:index}

The \emph{CoT Mediation Index} is a bounded score in $[0,1]$ that quantifies causal attribution uniquely to the CoT token positions:
\begin{equation}
\mathrm{CMI}_\ell
\;=\;
\begin{cases}
0, & \Delta_{\mathrm{cot},\ell} + \Delta_{\mathrm{ctrl},\ell} < \tau_{\mathrm{drop}},\\
\dfrac{\max\!\bigl(0,\, \Delta_{\mathrm{cot},\ell} - \Delta_{\mathrm{ctrl},\ell}\bigr)}{\max\!\bigl(\Delta_{\mathrm{cot},\ell} + \Delta_{\mathrm{ctrl},\ell},\, \tau_{\mathrm{den}}\bigr)}, & \text{otherwise,}
\end{cases}
\end{equation}
where $\tau_{\mathrm{drop}}$ is a drop floor and $\tau_{\mathrm{den}}$ is a denominator floor.

If $\Delta_{\mathrm{cot},\ell}+\Delta_{\mathrm{ctrl},\ell}$ is tiny, the estimate is dominated by numerical noise and sampling variance, so we set $\mathrm{CMI}_\ell=0$ to avoid spurious causal effects. Also, even when drops are non-zero, very small totals can make the ratio unstable; using $\max(\Delta_{\mathrm{cot},\ell}+\Delta_{\mathrm{ctrl},\ell},\tau_{\mathrm{den}})$ prevents tiny denominators from inflating $\mathrm{CMI}_\ell$.
Subtracting $\Delta_{\mathrm{ctrl},\ell}$ isolates CoT-specific effects, while dividing by a floored sum stabilizes the ratio, and the hard-zero condition avoids noise when both drops are tiny. We define Bypass as $1 - \mathrm{CMI}_\ell$. Low CMI indicates that the model's decision is weakly coupled to the explicit CoT text under these interventions. We treat peaks (or bands) in the CMI profile as candidate ``reasoning windows'' where CoT-token states seem most causally important.
This is an analysis heuristic.

\subsection{Behavioral proxies of faithfulness}
\label{sec:method:behavior}

To contrast mechanistic perspectives, we also report behavioral proxies commonly used to assess reasoning faithfulness (e.g., With-CoT versus No-CoT accuracy and, optionally, self-consistency). These metrics capture output-level differences but do not reveal whether internal computation causally depends on CoT. Comparing behavioral signals with CMI profiles highlights cases where apparent CoT benefits do not correspond to increased mechanistic reliance on CoT. Additional behavioral baseline details are provided in Appendix~\ref{app:behavior}.

\section{Results}
\label{sec:results}

We apply our causal intervention framework to multiple models and tasks to examine where and whether models mechanistically rely on chain-of-thought. We focus on three main questions: (i) where CoT Mediation Index is localized in depth, (ii) how this varies across models, and (iii) when models exhibit bypass regimes.

\subsection{Localization of CMI and Comparison Across Models}
\label{sec:results:localization}

We first examine representative models across families and scales (e.g., Qwen3-0.6B, Phi-4, Phi-4-Mini-Reasoning) and analyze their layerwise CMI profiles. Across many models, we observe that CMI is depth-localized, i.e. CMI peaks are concentrated within a limited range of layers rather than being uniformly distributed across the network in most models. We count a layer as active if the CMI, averaged across prompts, is greater than zero. For the \% Active Layers, we find percent ratio of the average number of active layers per prompt by the number of layers in that model. Table \ref{tab:cross_model_density} shows CMI distribution variation across models.

\begin{table}[h]
\centering
\caption{Cross-model summary of CoT-specific causal mediation. We report mean layerwise CMI, the total number of layers, and which layers are CoT-active (Average CMI$_\ell>0$) . \% Active Layers is computed as $100\times$ (average number of active layers per prompt) divided by the model's layer count. All models are given as per their HuggingFace names.}
\label{tab:cross_model_density}
\vspace{0.5em}
\begin{tabular}{@{}lcccc@{}}
\hline
\textbf{Model} & \textbf{Mean CMI} & \textbf{Layers} & \textbf{CMI-active layers} & \textbf{\% Active Layers} \\ \hline
Phi-mini-MoE-instruct & 0.1230 & 32 & [0, 31] & 27.03\% \\
Phi-4-mini-reasoning  & 0.0820 & 32 & 1, 2, 4, 8, 10, 12, [14, 16], [21, 31] & 12.66\% \\
Qwen3-1.7B                 & 0.0555 & 28 & [0, 8], 26, 27 & 8.21\% \\
Phi-3.5-mini-instruct & 0.0452 & 36 & [2, 6], [8, 14], [25, 29], [33, 35] & 5.14\% \\
phi-2                 & 0.0107 & 32 & [29, 31] & 2.66\% \\
Qwen3-0.6B                 & 0.0107 & 32 & [29, 31] & 2.66\% \\
DialoGPT-large & 0.0137 & 36 & [0, 2] & 1.94\% \\
phi-1\_5              & 0.0092 & 24 & 21, 22 & 1.25\% \\
phi-4                 & 0.0065 & 40 & [3, 5], 38 & 0.75\% \\
Qwen3-8B                   & 0.0014 & 36 & 33 & 0.14\% \\
Qwen3-4B                   & 0.0000 & 36 & -- & 0.00\% \\ \hline
\end{tabular}
\end{table}

\begin{figure}[h]
  \centering
  \begin{minipage}{0.48\linewidth}
    \centering
    \textbf{(a)} DialoGPT-large\\
    \vspace{0.25em}
    \includegraphics[width=\linewidth]{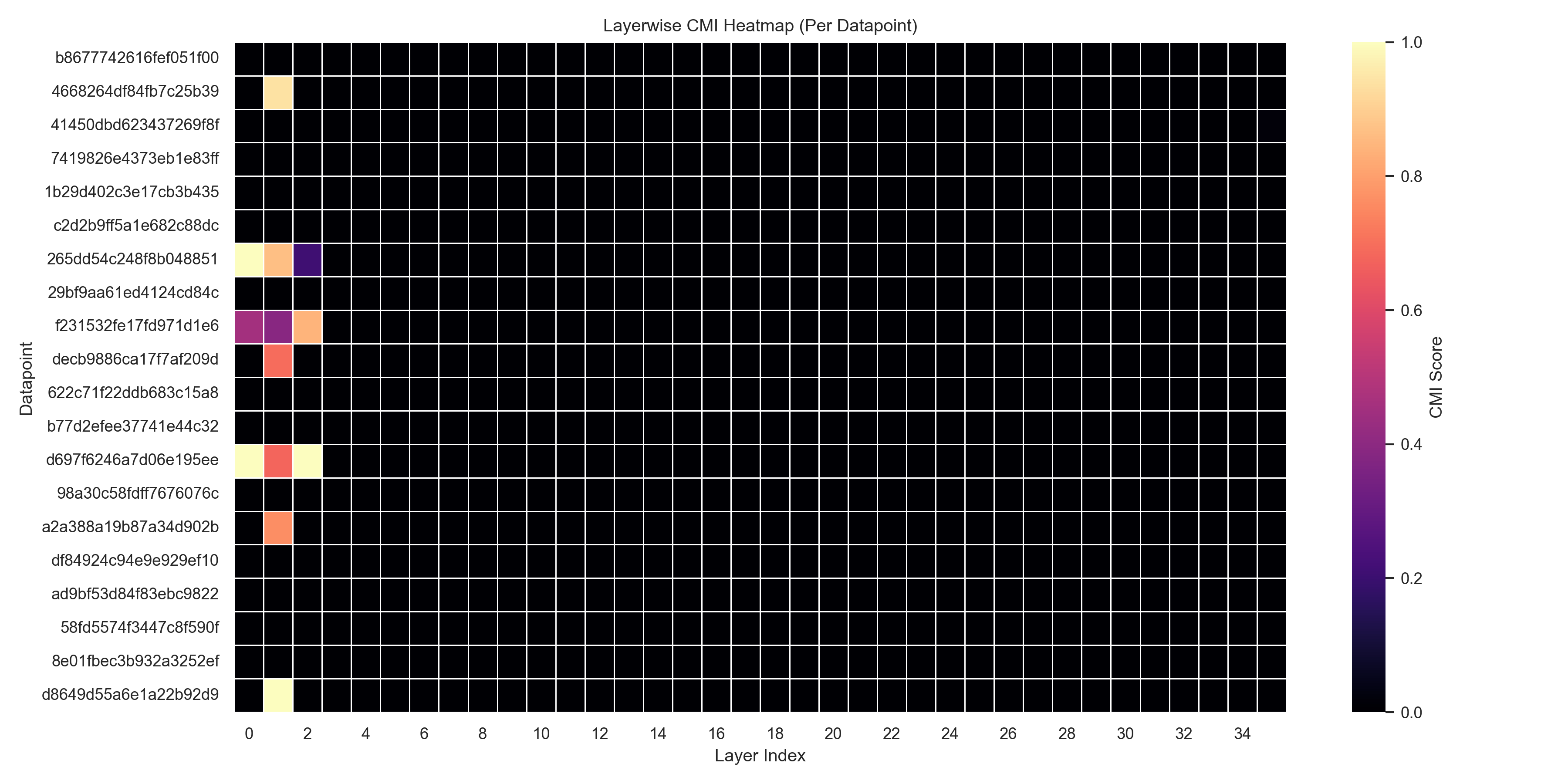}
  \end{minipage}
  \hfill
  \begin{minipage}{0.48\linewidth}
    \centering
    \textbf{(b)} Qwen3-0.6B\\
    \vspace{0.25em}
    \includegraphics[width=\linewidth]{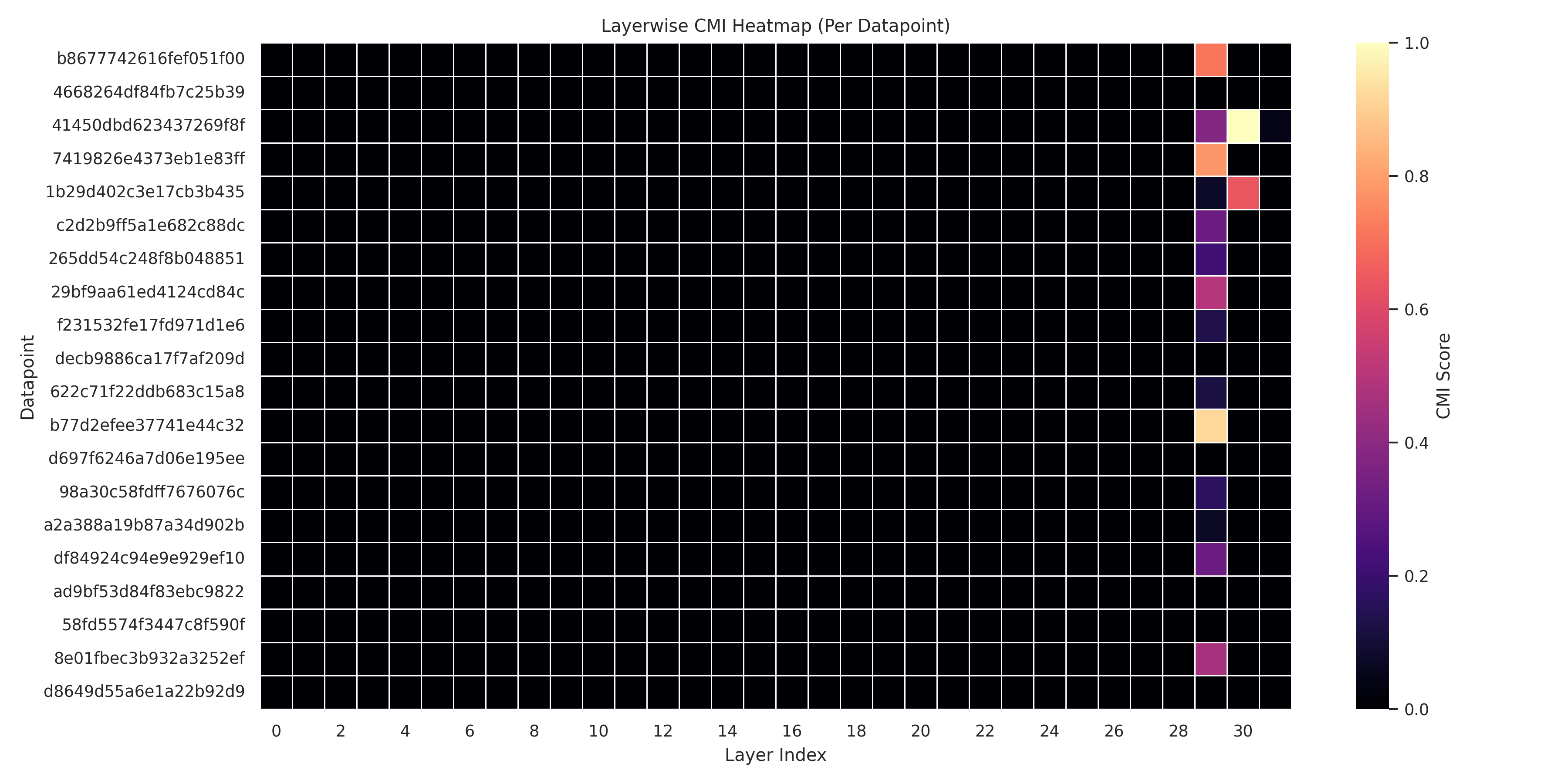}
  \end{minipage}
  \caption{Image (a) shows the CMI distribution for DialoGPT-large across 20 prompts in the form of a heatmap across layers. Image (b) shows the same for Qwen3-0.6B. The strings to the left of each graph indicate the qid of the prompt used in StrategyQA.}
  \label{fig:localization_side_by_side}
\end{figure}

Across models, we observe distinct \emph{routing regimes}. Some models exhibit strongly localized peaks, with most CoT-specific influence concentrated in a small set of deeper layers. Others show more distributed, moderate CMI across a broader range of layers, suggesting a more diffuse integration of reasoning signals. This cross-model diversity indicates that the internal use of CoT is not uniform across architectures or scales. The following experiments are run on the StrategyQA dataset (\cite{geva2021didaristotleuselaptop}). First, we highlight that different models follow CoT tokens at different layers, even given similar model sizes and number of layers. As can be seen in Figure \ref{fig:localization_side_by_side}, DialoGPT-large follows CoT in its early layers, while Qwen3-0.6B does so in its late layers. This indicates the dependence of CoT routing on model architecture.

\begin{figure}[h]
  \centering
  \begin{minipage}{0.48\linewidth}
    \centering
    \textbf{(a)} phi-4\\
    \vspace{0.25em}
    \includegraphics[width=\linewidth]{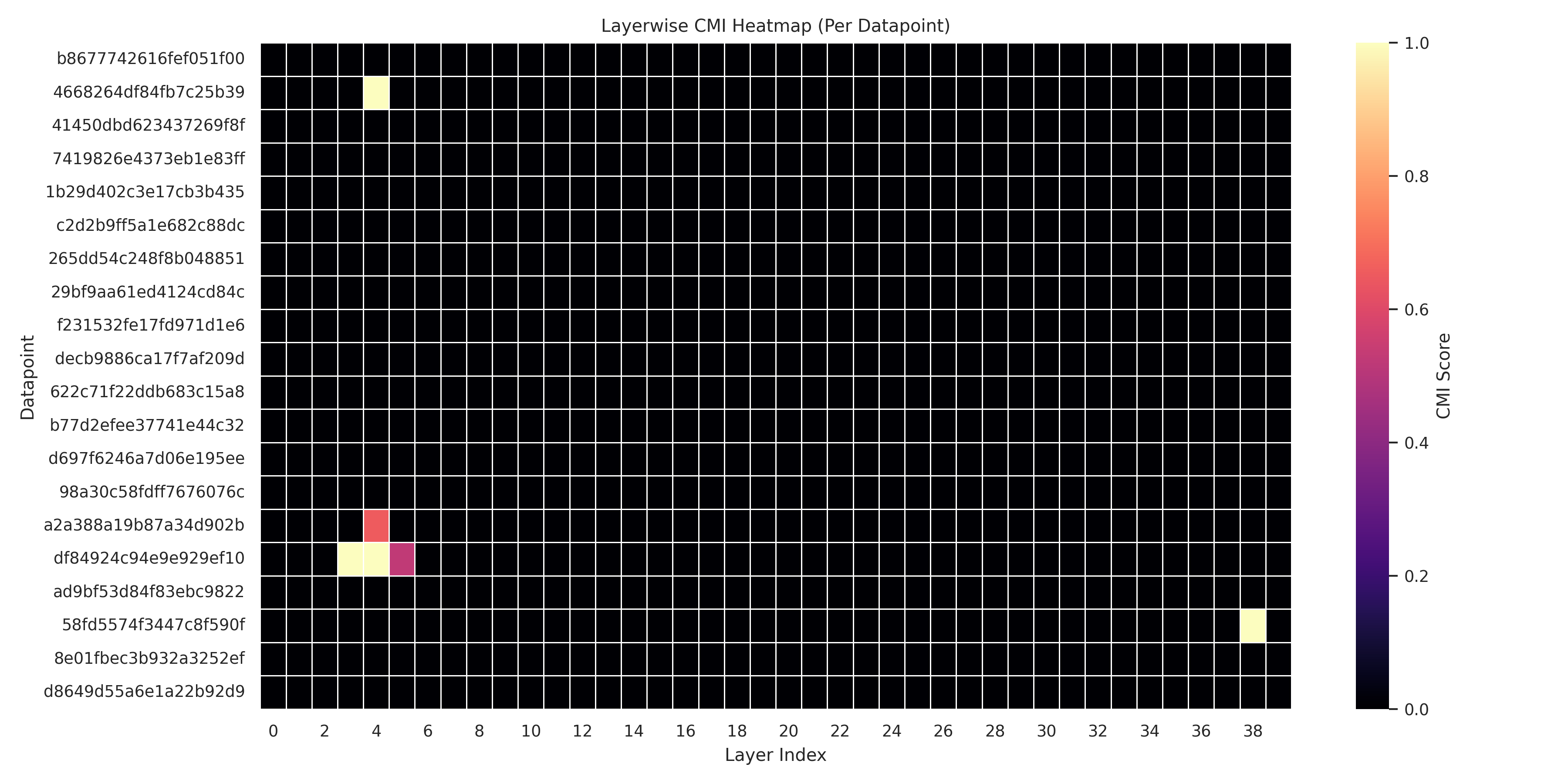}
  \end{minipage}
  \hfill
  \begin{minipage}{0.48\linewidth}
    \centering
    \textbf{(b)} Phi-4-mini-reasoning\\
    \vspace{0.25em}
    \includegraphics[width=\linewidth]{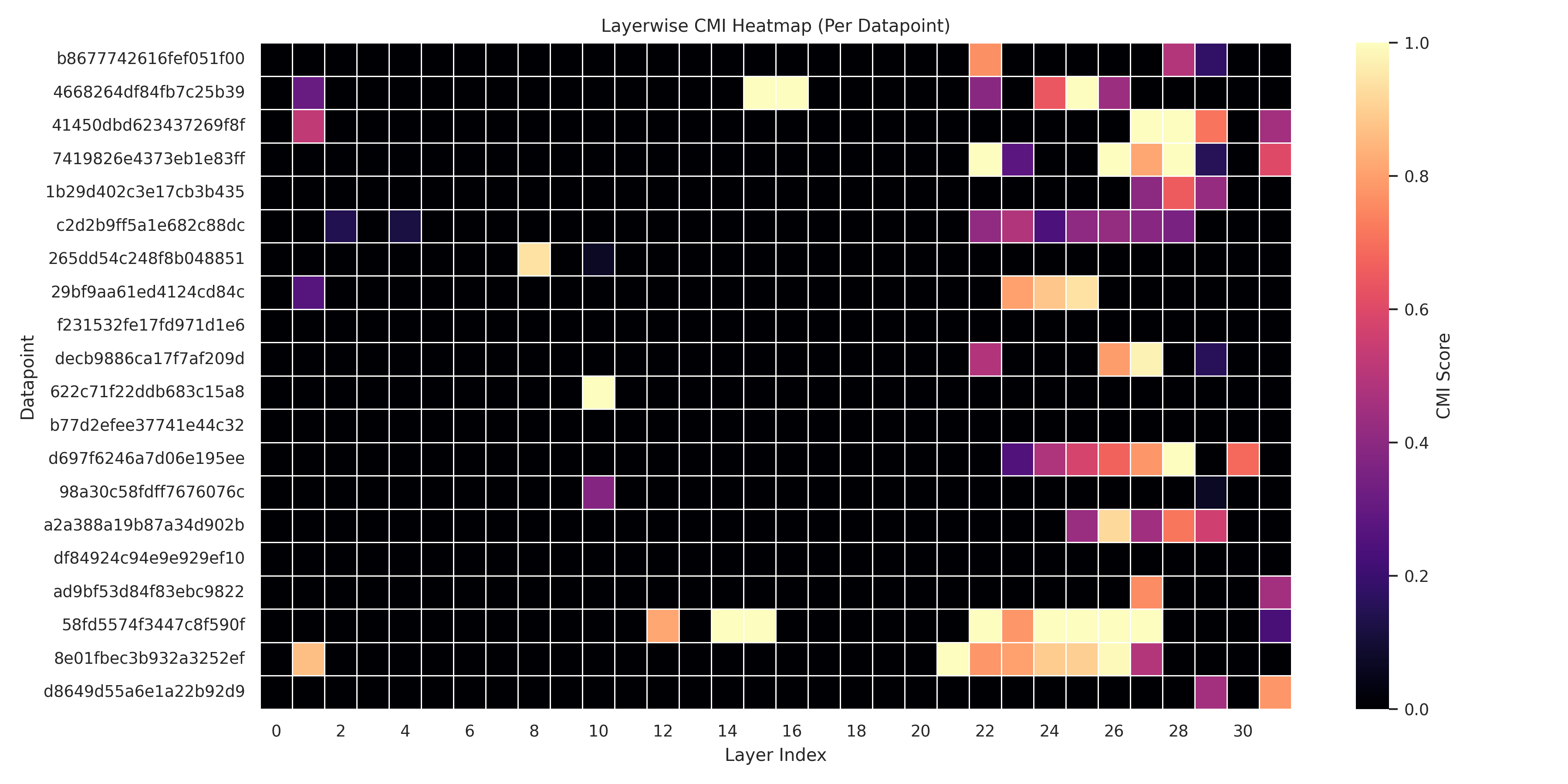}
  \end{minipage}
  \caption{Image (a) shows the CMI distribution for phi-4 across 20 prompts in the form of a heatmap across layers. Image (b) shows the same for Phi-4-mini-reasoning. The strings to the left of each graph indicate the qid of the prompt used in StrategyQA.}
  \label{fig:phicomparison}
\end{figure}

We also point out that simply scaling models without specifically training them for reasoning does not lead to more faithfulness to CoT. This can be seen in Figure \ref{fig:phicomparison}, where phi-4 has significantly lower CMI activation as compared to Phi-4-mini-reasoning, even though the former is nearly four times as large. We theorise this to be due to the latter being trained specifically for reasoning, while the former is simply a pre-trained model.

Furthermore, we notice that Mixture-of-Experts (MoE) models have more distributed CMI profiles, as can be seen in Figure~\ref{fig:strategyqa_microsoftphiMOEminiinstruct}.
A plausible explanation for this is that their computation is routed rather than uniform across depth. In dense transformers, shared weights may encourage narrow ``reasoning windows'' where CoT information is integrated at specific layers and then compressed into a latent state. By contrast, MoE layers dynamically select different experts for different tokens, which could allow CoT-aligned representations to be processed by specialized sub-networks at multiple depths. This routing-based structure may reduce the need for a single integration bottleneck and instead lead to repeated, shallow incorporation of CoT information across layers, yielding higher causal density but lower single-layer peaks. Under this hypothesis, MoE models would implement a more distributed form of reasoning, where influence accumulates through conditional expert pathways rather than a single depth-localized circuit.

\subsection{Causal Influence TruthfulQA and GSM8K}
\label{sec:results:truthfulqa}

To investigate whether chain-of-thought can override common misconceptions, we apply our causal bypass monitor to the TruthfulQA dataset~\citep{lin2022truthfulqameasuringmodelsmimic}, which targets human-like falsehoods that models frequently mimic due to their high training frequency. We find a near-total bypass regime across almost all instances, with $\mathrm{CMI} \approx 0$, possibly indicating little CoT-specific causal mediation. We also observe that baseline log-probability can substantially favor the myth answer in some instances (e.g., \texttt{tqa\_4}, \texttt{tqa\_7}). In rare cases such as \texttt{tqa\_1}, CMI is slightly higher for the myth answer than the truthful one. Full instance-level metrics (Table~\ref{tab:truthful_qa_results}) and detailed analysis are provided in Appendix~\ref{app:truthfulqa}. We also test on GSM8K~\citep{cobbe2021trainingverifierssolvemath}, a collection of linguistically diverse grade-school math word problems (see Appendix~\ref{app:gsm8k}), where we observe that instances that are computationally expensive for the model (e.g., multi-step arithmetic needing longer intermediate computation) tend to exhibit higher CMI, whereas low-computation cases tend to have low CMI.

% \subsection{Evidence for bypass regimes}
% \label{sec:results:bypass}

% We also identify cases where models produce plausible CoT behaviorally but show little additional mechanistic reliance on it. In these \emph{bypass regimes}, CMI values are low or near-zero across most layers, indicating that patching CoT-related representations does not substantially degrade performance beyond Control interventions.

% Importantly, some of these models still show behavioral improvements when prompted with CoT. This divergence between behavioral metrics and mechanistic evidence suggests that visible reasoning traces can be weakly coupled to internal computation. In such cases, CoT may function more as a stylistic or training-aligned output pattern rather than a faithful representation of the model’s internal reasoning pathway.

\section{Discussion}
\label{sec:discussion}

We directly measure where CoT-token representations causally affect answers. Across models, CoT influence is often concentrated in a narrow band of layers (a ``reasoning window''), with limited dependence elsewhere unless the model is specifically trained for reasoning. We also observe \textbf{bypass regimes} where models generate plausible CoT but show little mechanistic reliance on it, suggesting decisions can proceed through latent pathways weakly coupled to the emitted reasoning text. Thus, visible explanations can diverge from the internal processes that determine outputs.Cross-model comparisons indicate that CoT faithfulness is \textbf{not a monotonic function} of scale. Larger or more capable models do not necessarily show stronger CoT Mediation Index, whereas models trained explicitly for reasoning can exhibit more pronounced and structured CMI profiles. Additionally, Mixture-of-Experts architectures tend to show more distributed mediation, suggesting that routing-based designs may support more diffuse integration of reasoning signals. Together, these findings show that behavioral gains under CoT prompting do not guarantee mechanistic reliance on the reasoning trace. Causal, layerwise audits therefore help distinguish when CoT is mechanistically meaningful versus when it is primarily stylistic or post-hoc.

\section{Conclusion}
\label{sec:conclusion}

We introduced a causal, layerwise framework for auditing whether large language models mechanistically rely on chain-of-thought (CoT) reasoning. Using the CoT Mediation Index (CMI), we showed that visible reasoning traces and internal computation often diverge: CoT influence is frequently depth-localized and, in many cases, weak, revealing \emph{bypass regimes} where models generate fluent explanations without strongly depending on them internally. These results suggest that substantial reasoning may occur in \emph{implicit latent representations}, with CoT often serving as an imperfect summary rather than the primary computational pathway. Overall, this framework helps distinguish when models are ``thinking in text'' versus ``thinking in latent space,'' improving reasoning transparency and interpretability analyses.

\bibliography{references}

@misc{greenblatt2024alignmentfaking,
  author        = {Greenblatt, R. and Denison, C. and Wright, B. and Roger, F. and MacDiarmid, M. and Marks, S. and Treutlein, J. and Belonax, T. and Chen, J. and Duvenaud, D. and Khan, A. and Michael, J. and Mindermann, S. and Perez, E. and Petrini, L. and Uesato, J. and Kaplan, J. and Shlegeris, B. and Bowman, S. R. and Hubinger, E.},
  title         = {Alignment Faking in Large Language Models},
  year          = {2024},
  archivePrefix = {arXiv},
  eprint        = {2412.14093},
  url           = {https://arxiv.org/abs/2412.14093},
  doi           = {10.48550/arXiv.2412.14093}
}

@misc{baker2025monitoringreasoning,
  author        = {Baker, B. and Huizinga, J. and Farhi, D. and Gao, L. and Dou, Z. and Guan, M. Y. and Madry, A. and Zaremba, W. and Pachocki, J.},
  title         = {Monitoring Reasoning Models for Misbehavior and the Risks of Promoting Obfuscation},
  year          = {2025},
  archivePrefix = {arXiv},
  eprint        = {2503.11926},
  url           = {https://arxiv.org/abs/2503.11926},
  doi           = {10.48550/arXiv.2503.11926}
}

@misc{wang2025whenthinkingllmslie,
  author        = {Wang, K. and Zhang, Y. and Sun, M.},
  title         = {When Thinking LLMs Lie: Unveiling the Strategic Deception in Representations of Reasoning Models},
  year          = {2025},
  archivePrefix = {arXiv},
  eprint        = {2506.04909},
  url           = {https://arxiv.org/abs/2506.04909},
  doi           = {10.48550/arXiv.2506.04909}
}

@misc{sharma2023sycophancy,
  author        = {Sharma, M. and Tong, M. and Korbak, T. and Duvenaud, D. and Askell, A. and Bowman, S. R. and Cheng, N. and Durmus, E. and Hatfield-Dodds, Z. and Johnston, S. R. and Kravec, S. and Maxwell, T. and McCandlish, S. and Ndousse, K. and Rausch, O. and Schiefer, N. and Yan, D. and Zhang, M. and Perez, E.},
  title         = {Towards Understanding Sycophancy in Language Models},
  year          = {2023},
  archivePrefix = {arXiv},
  eprint        = {2310.13548},
  url           = {https://arxiv.org/abs/2310.13548},
  doi           = {10.48550/arXiv.2310.13548}
}

@misc{yee2024dissociation,
  author        = {Yee, Evelyn and Li, Alice and Tang, Chenyu and Jung, Yeon Ho and Paturi, Ramamohan and Bergen, Leon},
  title         = {Dissociation of Faithful and Unfaithful Reasoning in {LLM}s},
  year          = {2024},
  archivePrefix = {arXiv},
  eprint        = {2405.15092},
  url           = {https://arxiv.org/abs/2405.15092},
  doi           = {10.48550/arXiv.2405.15092}
}

@misc{chen2025reasoning,
  author        = {Chen, Yanda and Benton, Joe and Radhakrishnan, Ansh and Uesato, Jonathan and Denison, Carson and Schulman, John and Somani, Arushi and Hase, Peter and Wagner, Misha and Roger, Fabien and Mikulik, Vlad and Bowman, Samuel R. and Leike, Jan and Kaplan, Jared and Perez, Ethan},
  title         = {Reasoning Models Don't Always Say What They Think},
  year          = {2025},
  archivePrefix = {arXiv},
  eprint        = {2505.05410},
  url           = {https://arxiv.org/abs/2505.05410},
  doi           = {10.48550/arXiv.2505.05410}
}

@misc{khalifa2026gaming,
  author        = {Khalifa, Muhammad and Logeswaran, Lajanugen and Kim, Jaekyeom and Sohn, Sungryull and Zhang, Yunxiang and Lee, Moontae and Peng, Hao and Wang, Lu and Lee, Honglak},
  title         = {Gaming the Judge: Unfaithful Chain-of-Thought Can Undermine Agent Evaluation},
  year          = {2026},
  archivePrefix = {arXiv},
  eprint        = {2601.14691},
  url           = {https://arxiv.org/abs/2601.14691},
  doi           = {10.48550/arXiv.2601.14691}
}

@misc{heimersheim2024activation,
  author        = {Heimersheim, Stefan and Nanda, Neel},
  title         = {How to Use and Interpret Activation Patching},
  year          = {2024},
  archivePrefix = {arXiv},
  eprint        = {2404.15255},
  url           = {https://arxiv.org/abs/2404.15255},
  doi           = {10.48550/arXiv.2404.15255}
}

@misc{zhang2023bestpractices,
  author        = {Zhang, Fred and Nanda, Neel},
  title         = {Towards Best Practices of Activation Patching in Language Models: Metrics and Methods},
  year          = {2023},
  archivePrefix = {arXiv},
  eprint        = {2309.16042},
  url           = {https://arxiv.org/abs/2309.16042},
  doi           = {10.48550/arXiv.2309.16042}
}

@misc{lin2022truthfulqameasuringmodelsmimic,
      title={TruthfulQA: Measuring How Models Mimic Human Falsehoods}, 
      author={Stephanie Lin and Jacob Hilton and Owain Evans},
      year={2022},
      eprint={2109.07958},
      archivePrefix={arXiv},
      primaryClass={cs.CL},
      url={https://arxiv.org/abs/2109.07958}, 
}

@misc{geva2021didaristotleuselaptop,
      title={Did Aristotle Use a Laptop? A Question Answering Benchmark with Implicit Reasoning Strategies}, 
      author={Mor Geva and Daniel Khashabi and Elad Segal and Tushar Khot and Dan Roth and Jonathan Berant},
      year={2021},
      eprint={2101.02235},
      archivePrefix={arXiv},
      primaryClass={cs.CL},
      url={https://arxiv.org/abs/2101.02235}, 
}

@misc{cobbe2021trainingverifierssolvemath,
      title={Training Verifiers to Solve Math Word Problems}, 
      author={Karl Cobbe and Vineet Kosaraju and Mohammad Bavarian and Mark Chen and Heewoo Jun and Lukasz Kaiser and Matthias Plappert and Jerry Tworek and Jacob Hilton and Reiichiro Nakano and Christopher Hesse and John Schulman},
      year={2021},
      eprint={2110.14168},
      archivePrefix={arXiv},
      primaryClass={cs.LG},
      url={https://arxiv.org/abs/2110.14168}, 
}
\bibliographystyle{iclr2026_conference}

\appendix

\section{Limitations and Future Work}
\label{sec:limitations}

\begin{enumerate}
\item \textbf{Intervention-based fragility.}  
Our causal analysis relies on activation patching, which can introduce distribution shift and model fragility unrelated to the targeted mechanism. Although we subtract matched control interventions to isolate CoT-specific effects, residual confounds may remain, especially in deeper layers where representations are highly entangled.

\item \textbf{Scope of tasks and prompts.}  
Our evaluation covers synthetic arithmetic/logic tasks and selected reasoning benchmarks, but does not exhaust the diversity of reasoning behaviors (e.g., long-horizon planning, tool use, multimodal reasoning). The observed routing regimes may vary across domains. Broader task coverage is needed to assess greater generality.

\item \textbf{Scale and compute constraints.}  
Layerwise causal intervention is computationally expensive, limiting the number of prompts, models, and ablations we can test. This constrains statistical power and breadth of analysis. Future work could develop scalable approximations or learned proxies for CMI that retain mechanistic interpretability.

\item \textbf{Behavioral baseline limitations.}  
Our surface-level behavioral signals are lightweight and interpretable but are not calibrated detectors of manipulation. They may produce false positives (benign meta-reasoning) and false negatives (subtle or obfuscated strategies). These signals should be treated as heuristic triage tools. Future work could train supervised faithfulness classifiers and test robustness to adversarial paraphrasing.

\item \textbf{Limits of causal interpretation.}  
Low CMI does not imply absence of reasoning; it indicates that answer-relevant computation is not mediated by CoT-aligned representations. Models may rely on latent pathways that are not text-aligned. Future work should pair our approach with circuit-level or latent-feature interpretability methods to better characterize these bypass pathways.

\item \textbf{Future directions.}  
Promising extensions include auditing process-supervised or reasoning-tuned models to test whether training increases genuine CoT reliance, linking CMI patterns to architectural features such as MoE routing, studying how routing regimes change with scale, and integrating mechanistic audits with behavioral monitoring for more robust evaluation in safety-critical settings.
\end{enumerate}

\section{Intervention Details}
\label{app:interventions}

\paragraph{Patch locations.}
Our intervention targets the \emph{token-level hidden states} at specific layers and token positions. For each example, we define a set of CoT token positions within the With-CoT prompt (identified by string matching and token alignment in the prompt). We then patch the \emph{hidden states at those positions} in the With-CoT forward pass using hidden states taken from the corresponding No-CoT run. Concretely:
\begin{itemize}
\item \textbf{Primary patch (CoT positions).} For a given layer (or layer span), we replace hidden states at the CoT token positions in the With-CoT run with the hidden states from the No-CoT run, at the same layer and token indices. This is the core causal intervention.
\item \textbf{Control patch (random non-CoT positions).} We select a matched-size random subset of \emph{non-CoT} token positions and perform the same kind of patching at those positions. This yields the control drop, which estimates the generic sensitivity to patching at that layer.
\end{itemize}
All patching is applied at the layer output (the hidden state tensor for that layer), and only at the specified token indices, leaving all other tokens untouched.

\paragraph{Implementation mechanics.}
The intervention uses a forward-patching mechanism that replaces hidden states at the chosen positions before subsequent layers are computed. The sequence of steps is:
\begin{enumerate}
\item Run the model on both the With-CoT and No-CoT inputs with hidden states enabled, producing hidden state tensors for each layer.
\item For a target layer (or a span of layers), build a \emph{patch tensor} by copying the No-CoT hidden states at the selected token positions and inserting them into the With-CoT hidden state tensor.
\item Use a context manager that patches model layers during the forward pass. Inside this context, recompute the answer log-probability for the With-CoT input with the patched states.
\item Compute the log-probability drop as the difference between the baseline (unpatched) log-probability and the patched log-probability, clipped at zero.
\end{enumerate}
This ensures a clean causal comparison: the only change between the baseline and patched computation is the hidden state content at the selected token positions and layer(s).

\paragraph{Intervention hyperparameters.}
We control patching and measurement with the following hyperparameters (as defined in the implementation):
\begin{itemize}
\item \textbf{Control samples.} For each layer/span, we draw multiple random control patches and average their effect. This reduces variance in the control estimate. (Default: \texttt{control\_samples=8} in test scripts.)
\item \textbf{CoT span definition.} CoT token positions are computed by locating the CoT span in the prompt and then mapping character positions to token indices via the tokenizer; if not found, we fall back to the last few tokens.
\item \textbf{CMI floors.} To avoid instability from tiny denominators, CMI uses a drop floor and a denominator floor:
\[
\texttt{CMI\_DROP\_FLOOR} = 1\times 10^{-4}, \quad
\texttt{CMI\_DENOM\_FLOOR} = 1\times 10^{-3}.
\]
If the total drop is below \texttt{CMI\_DROP\_FLOOR}, CMI is set to zero.
\item \textbf{Additional reporting floors.} We also define \texttt{CMI\_BASE\_FLOOR} = $10^{-2}$ for stability checks and reporting.
\item \textbf{Generation settings.} When generating auxiliary CoT text for span-based patching, we use \texttt{max\_new\_tokens=80}, \texttt{temperature=0.7}, and \texttt{top\_p=0.9}. (These parameters only affect synthetic CoT generation, not the primary intervention on fixed dataset prompts.)
\end{itemize}

\paragraph{Ablations and robustness checks.}
We include several ablations to test whether CMI reflects genuine CoT-specific dependence rather than patching artifacts:
\begin{itemize}
\item \textbf{Placebo patching.} Replace CoT hidden states with random noise (or randomized states) instead of No-CoT activations and recompute CMI. This yields \emph{Placebo CMI}, which should be low if CMI is CoT-specific.
\item \textbf{Boundary sensitivity.} Expand or shrink the CoT span by one token on each side and measure changes in CMI. This yields a relative sensitivity metric that flags boundary-fragile results. This can be computed using layer-wise CMI values.
\end{itemize}
These ablations are implemented in the same framework and share the same patching and scoring infrastructure.

\paragraph{Reporting conventions.}
We report per-layer (or per-span) \texttt{cot\_drop}, \texttt{control\_drop}, \texttt{CMI}, and \texttt{Bypass} scores. When plotting or summarizing, we either (i) average CMI across layer spans or (ii) show the full layerwise profile $\{\mathrm{CMI}_\ell\}$ to identify localized reasoning windows.
This combination captures both the global tendency to route through CoT and the specific depths at which CoT becomes causally active.

\clearpage
\section{StrategyQA Results}
\label{app:strategyqa}

We provide additional layerwise intervention visualizations for StrategyQA. For each model, we show (left) the CoT-drop profile across depth and (right) the corresponding layerwise CMI heatmap.

\begin{figure}[h]
  \centering
  \begin{minipage}{0.49\linewidth}
    \centering
    \textbf{(a)} CoT drop\\
    \vspace{0.25em}
    \IfFileExists{modelsrun/new_plot_layer_drop_means_microsoftphiMOEminiinstruct.png}{%
      \includegraphics[width=\linewidth]{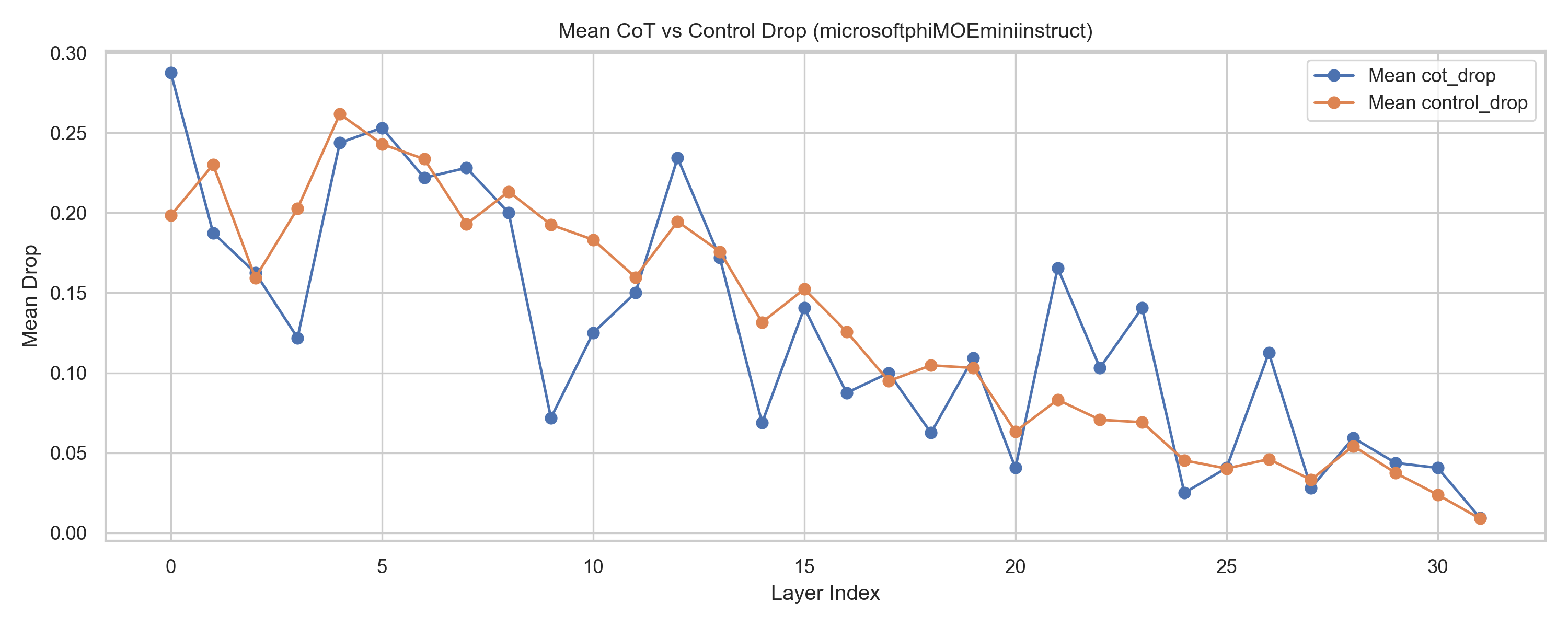}%
    }{\fbox{\rule[-.5cm]{0cm}{3.0cm}\rule[-.5cm]{\linewidth}{0cm}}}
  \end{minipage}
  \hfill
  \begin{minipage}{0.49\linewidth}
    \centering
    \textbf{(b)} CMI heatmap\\
    \vspace{0.25em}
    \IfFileExists{modelsrun/new_plot_layer_heatmap_microsoftphiMOEminiinstruct.png}{%
      \includegraphics[width=\linewidth]{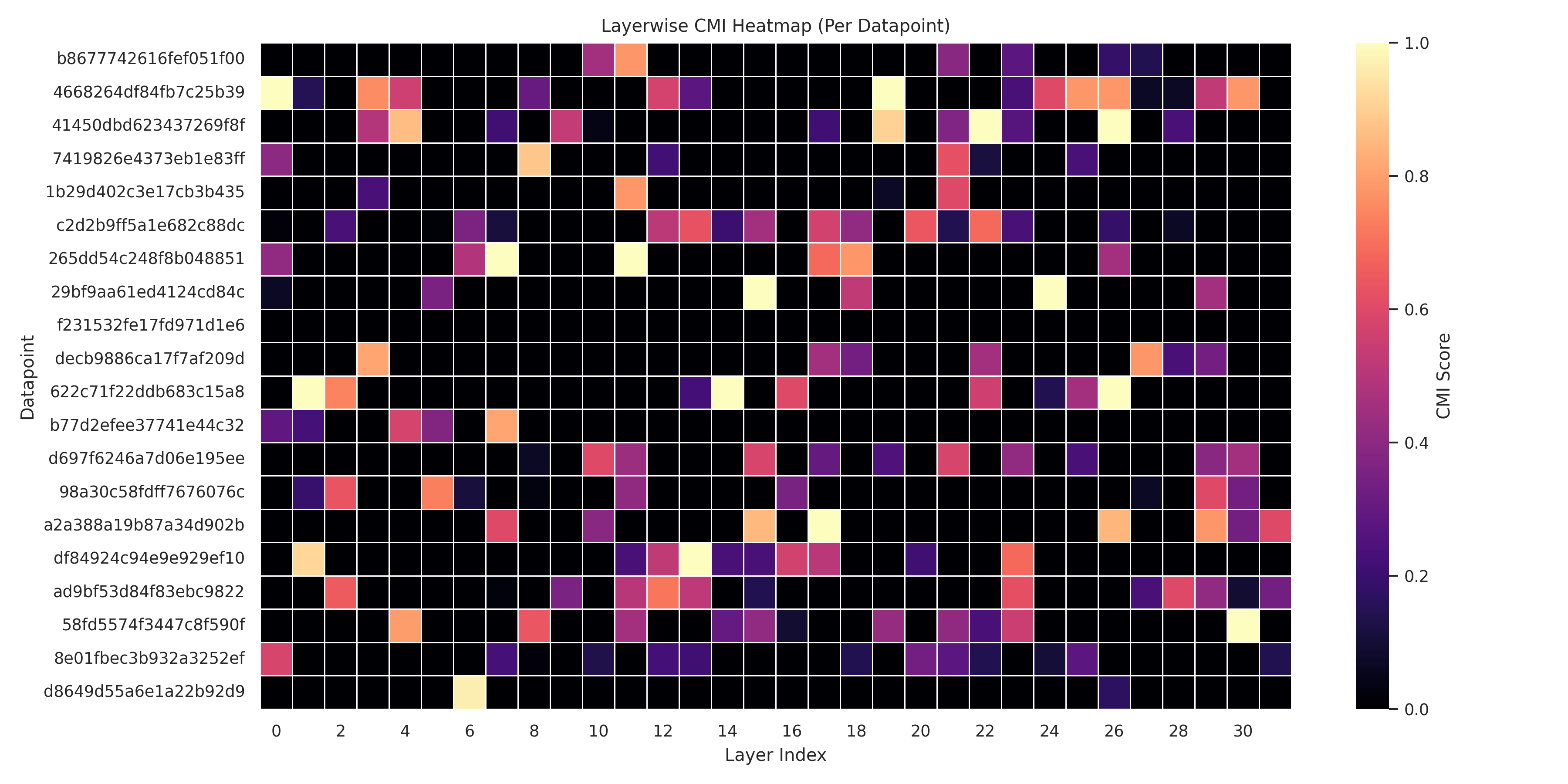}%
    }{\fbox{\rule[-.5cm]{0cm}{3.0cm}\rule[-.5cm]{\linewidth}{0cm}}}
  \end{minipage}
  \caption{StrategyQA layerwise CoT mediation for Phi-mini-MoE-instruct.}
  \label{fig:strategyqa_microsoftphiMOEminiinstruct}
\end{figure}

\begin{figure}[h]
  \centering
  \begin{minipage}{0.49\linewidth}
    \centering
    \textbf{(a)} CoT drop\\
    \vspace{0.25em}
    \IfFileExists{modelsrun/new_plot_layer_drop_means_microsoftphimini4reasoning.png}{%
      \includegraphics[width=\linewidth]{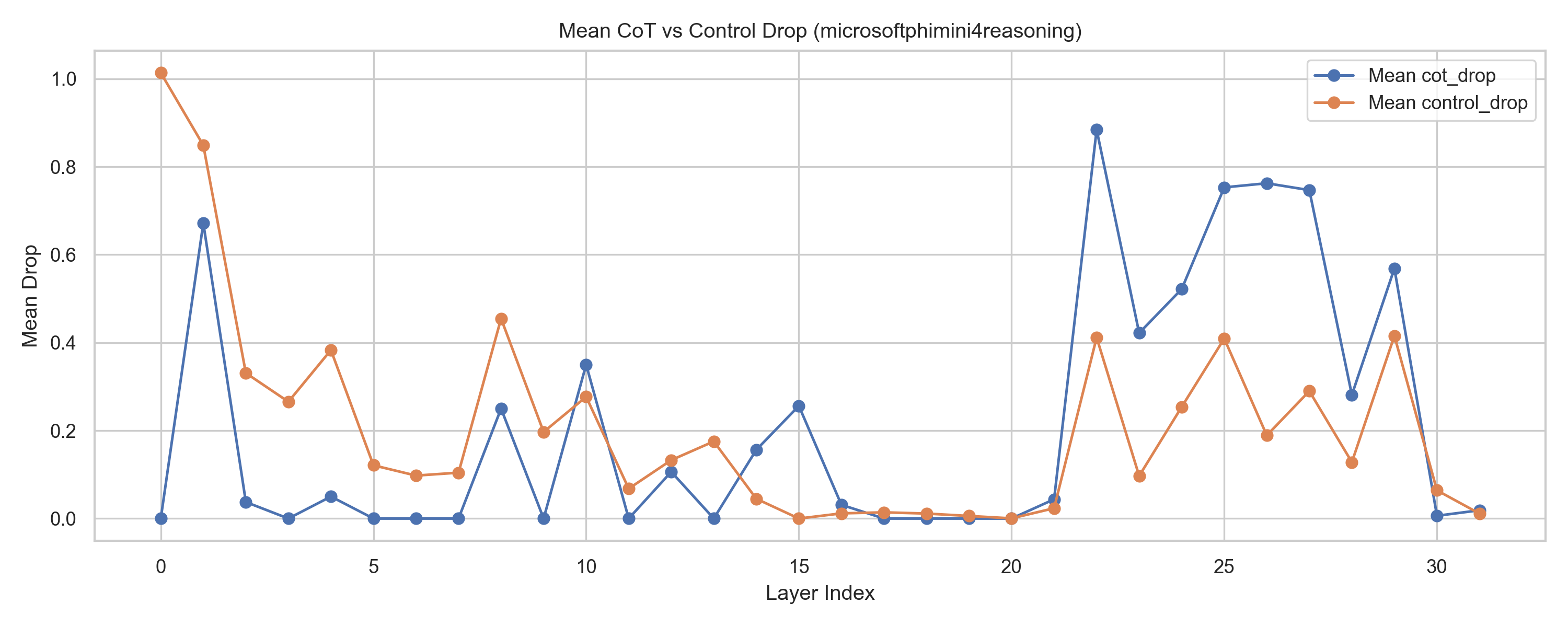}%
    }{\fbox{\rule[-.5cm]{0cm}{3.0cm}\rule[-.5cm]{\linewidth}{0cm}}}
  \end{minipage}
  \hfill
  \begin{minipage}{0.49\linewidth}
    \centering
    \textbf{(b)} CMI heatmap\\
    \vspace{0.25em}
    \IfFileExists{modelsrun/new_plot_layer_heatmap_microsoftphimini4reasoning.png}{%
      \includegraphics[width=\linewidth]{modelsrun/new_plot_layer_heatmap_microsoftphimini4reasoning.png}%
    }{\fbox{\rule[-.5cm]{0cm}{3.0cm}\rule[-.5cm]{\linewidth}{0cm}}}
  \end{minipage}
  \caption{StrategyQA layerwise CoT mediation for Phi-4-mini-reasoning.}
  \label{fig:strategyqa_microsoftphimini4reasoning}
\end{figure}

\begin{figure}[h]
  \centering
  \begin{minipage}{0.49\linewidth}
    \centering
    \textbf{(a)} CoT drop\\
    \vspace{0.25em}
    \IfFileExists{modelsrun/new_plot_layer_drop_means_microsoftphi3_5instruct.png}{%
      \includegraphics[width=\linewidth]{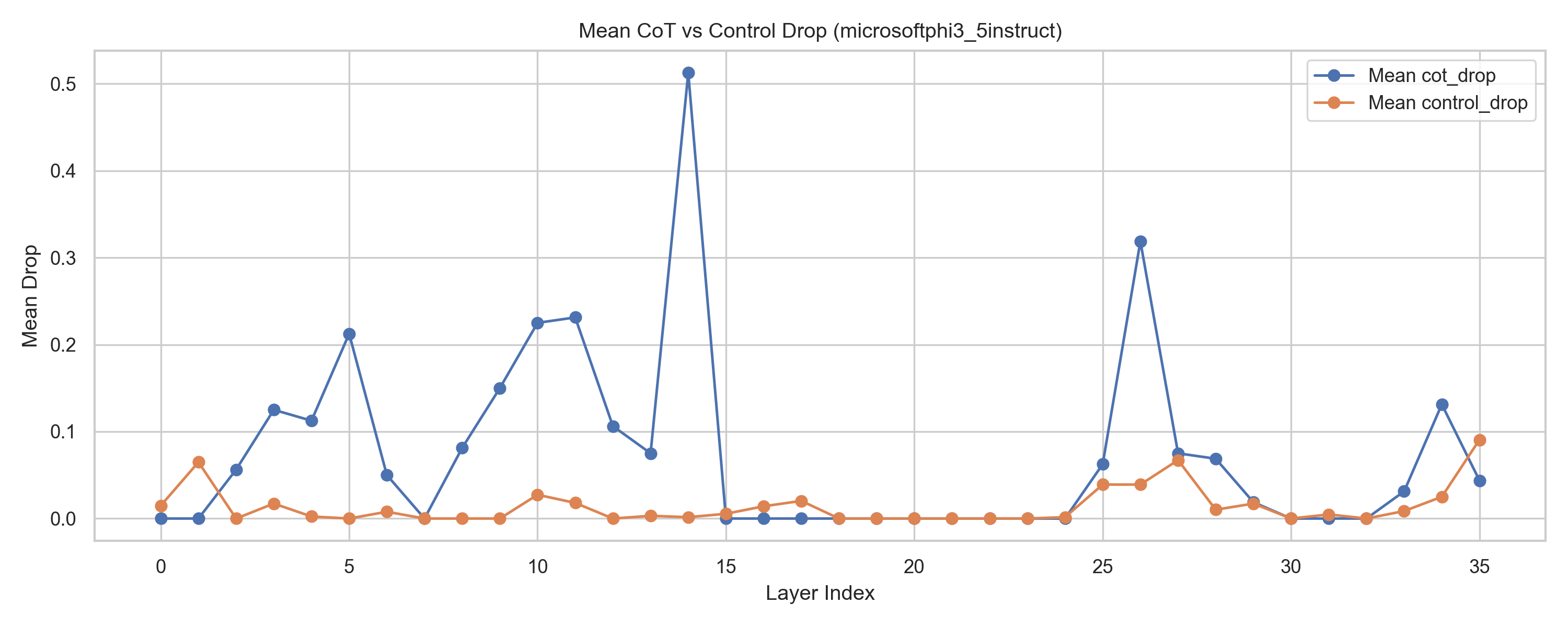}%
    }{\fbox{\rule[-.5cm]{0cm}{3.0cm}\rule[-.5cm]{\linewidth}{0cm}}}
  \end{minipage}
  \hfill
  \begin{minipage}{0.49\linewidth}
    \centering
    \textbf{(b)} CMI heatmap\\
    \vspace{0.25em}
    \IfFileExists{modelsrun/new_plot_layer_heatmap_microsoftphi3_5instruct.png}{%
      \includegraphics[width=\linewidth]{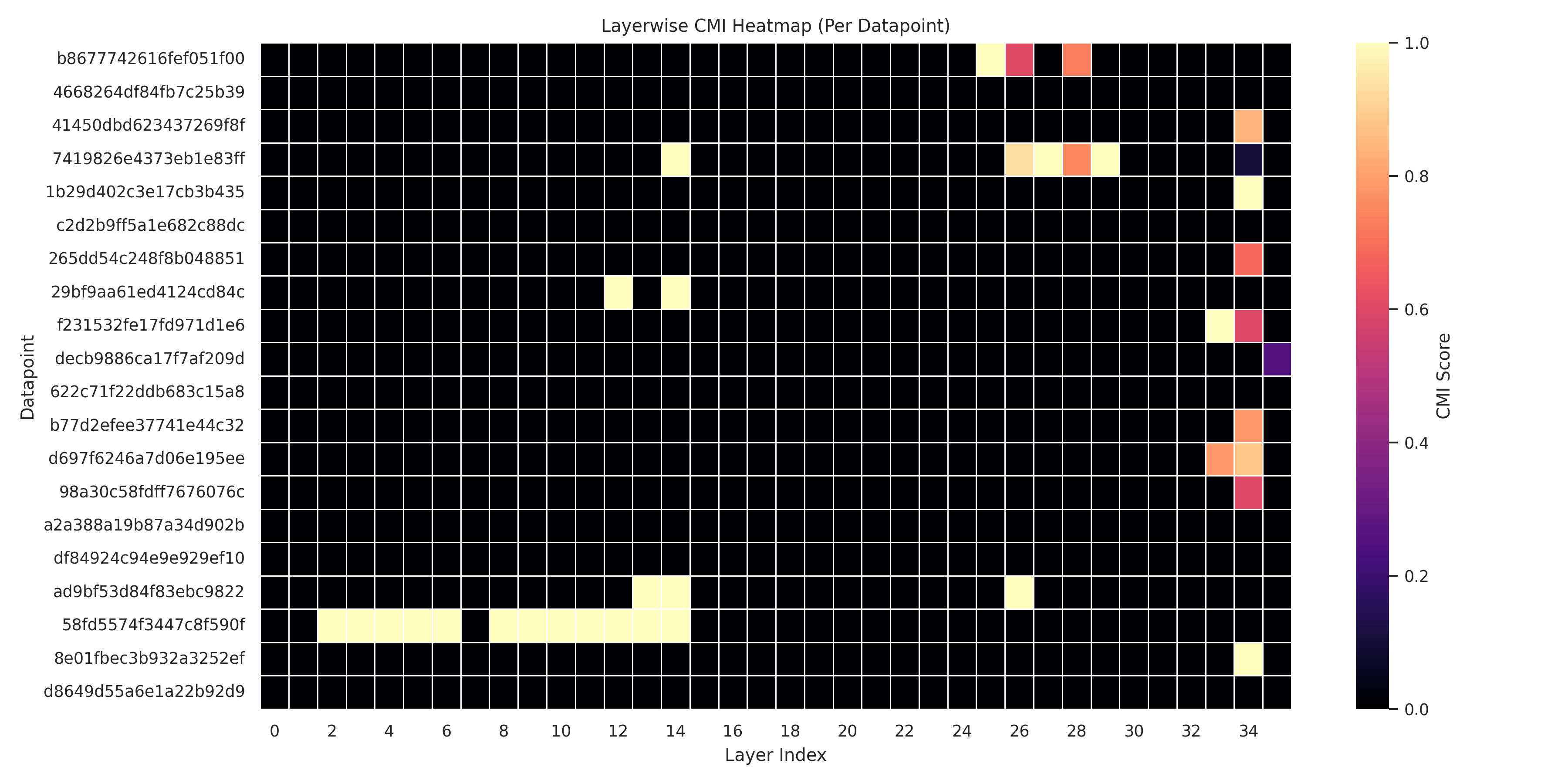}%
    }{\fbox{\rule[-.5cm]{0cm}{3.0cm}\rule[-.5cm]{\linewidth}{0cm}}}
  \end{minipage}
  \caption{StrategyQA layerwise CoT mediation for Phi-3.5-mini-instruct.}
  \label{fig:strategyqa_microsoftphi3_5instruct}
\end{figure}

\begin{figure}[h]
  \centering
  \begin{minipage}{0.49\linewidth}
    \centering
    \textbf{(a)} CoT drop\\
    \vspace{0.25em}
    \IfFileExists{modelsrun/new_plot_layer_drop_means_microsoftphi4.png}{%
      \includegraphics[width=\linewidth]{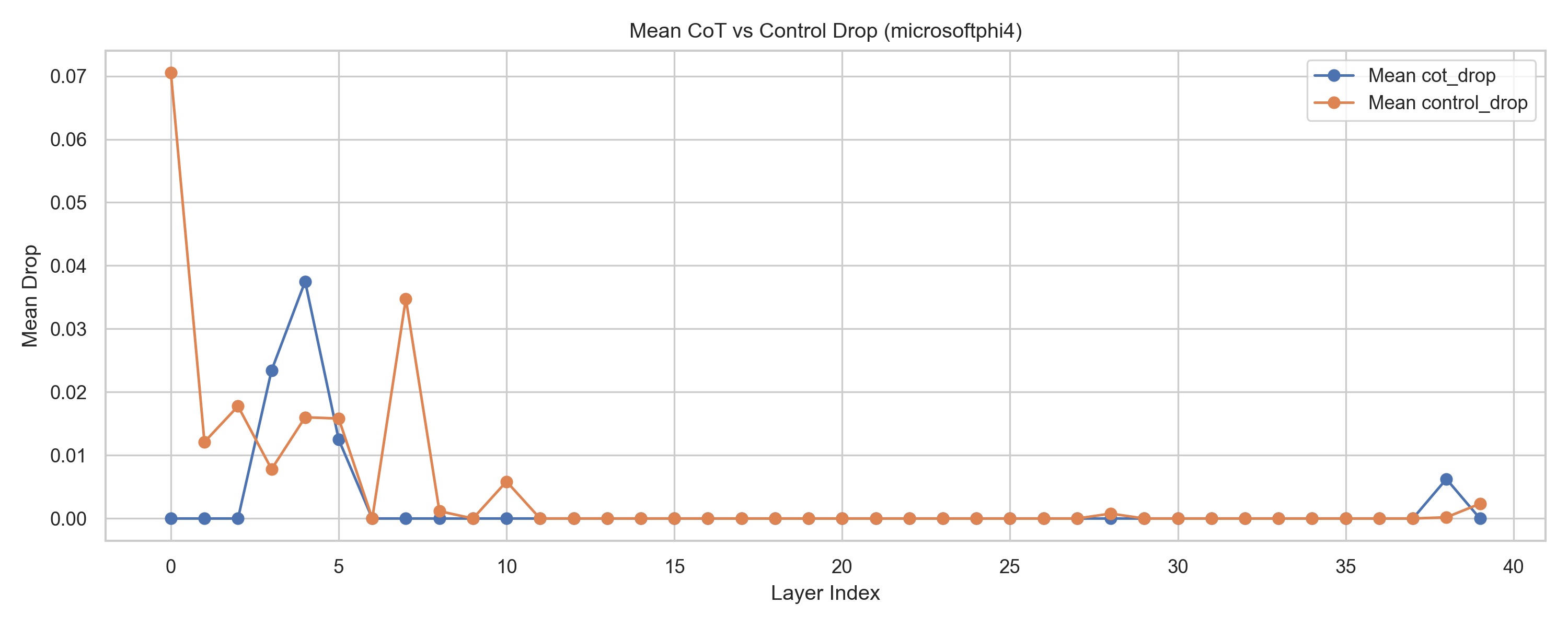}%
    }{\fbox{\rule[-.5cm]{0cm}{3.0cm}\rule[-.5cm]{\linewidth}{0cm}}}
  \end{minipage}
  \hfill
  \begin{minipage}{0.49\linewidth}
    \centering
    \textbf{(b)} CMI heatmap\\
    \vspace{0.25em}
    \IfFileExists{modelsrun/new_plot_layer_heatmap_microsoftphi4.png}{%
      \includegraphics[width=\linewidth]{modelsrun/new_plot_layer_heatmap_microsoftphi4.png}%
    }{\fbox{\rule[-.5cm]{0cm}{3.0cm}\rule[-.5cm]{\linewidth}{0cm}}}
  \end{minipage}
  \caption{StrategyQA layerwise CoT mediation for phi-4.}
  \label{fig:strategyqa_microsoftphi4}
\end{figure}

\begin{figure}[h]
  \centering
  \begin{minipage}{0.49\linewidth}
    \centering
    \textbf{(a)} CoT drop\\
    \vspace{0.25em}
    \IfFileExists{modelsrun/new_plot_layer_drop_means_microsoftphi2.png}{%
      \includegraphics[width=\linewidth]{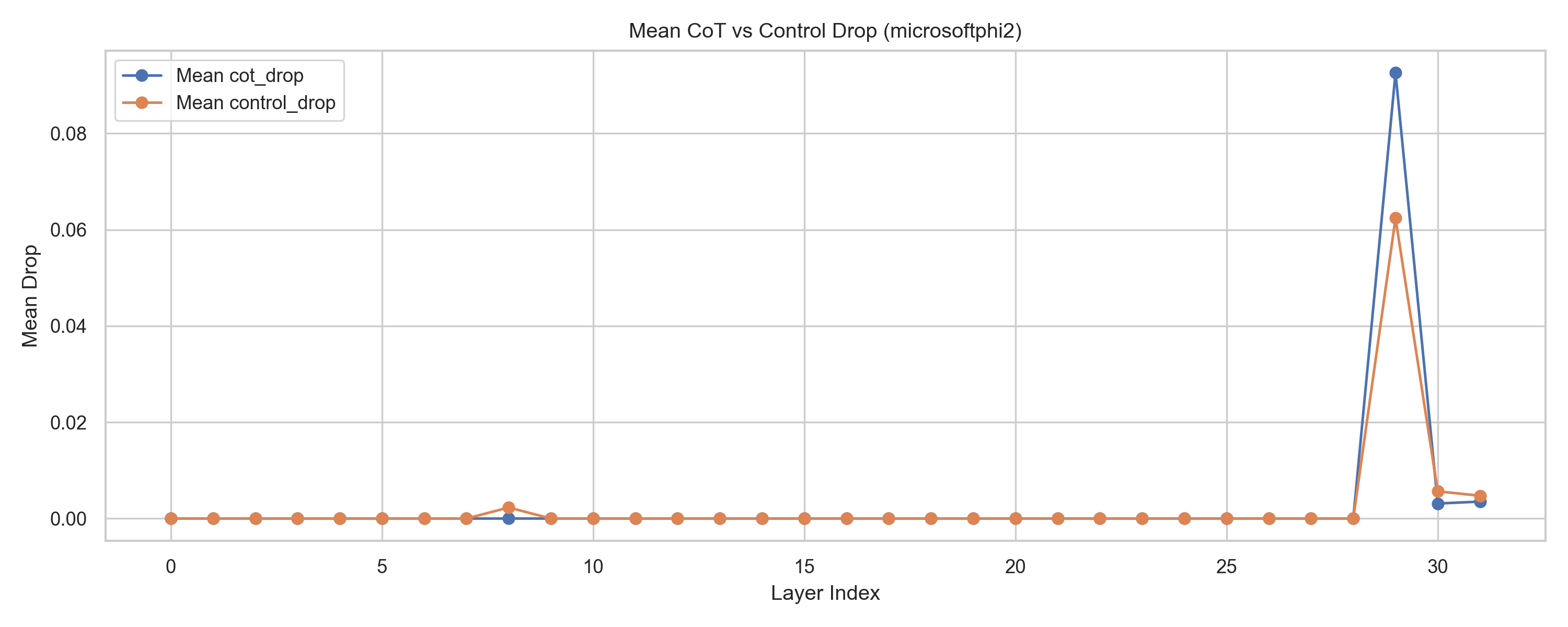}%
    }{\fbox{\rule[-.5cm]{0cm}{3.0cm}\rule[-.5cm]{\linewidth}{0cm}}}
  \end{minipage}
  \hfill
  \begin{minipage}{0.49\linewidth}
    \centering
    \textbf{(b)} CMI heatmap\\
    \vspace{0.25em}
    \IfFileExists{modelsrun/new_plot_layer_heatmap_microsoftphi2.png}{%
      \includegraphics[width=\linewidth]{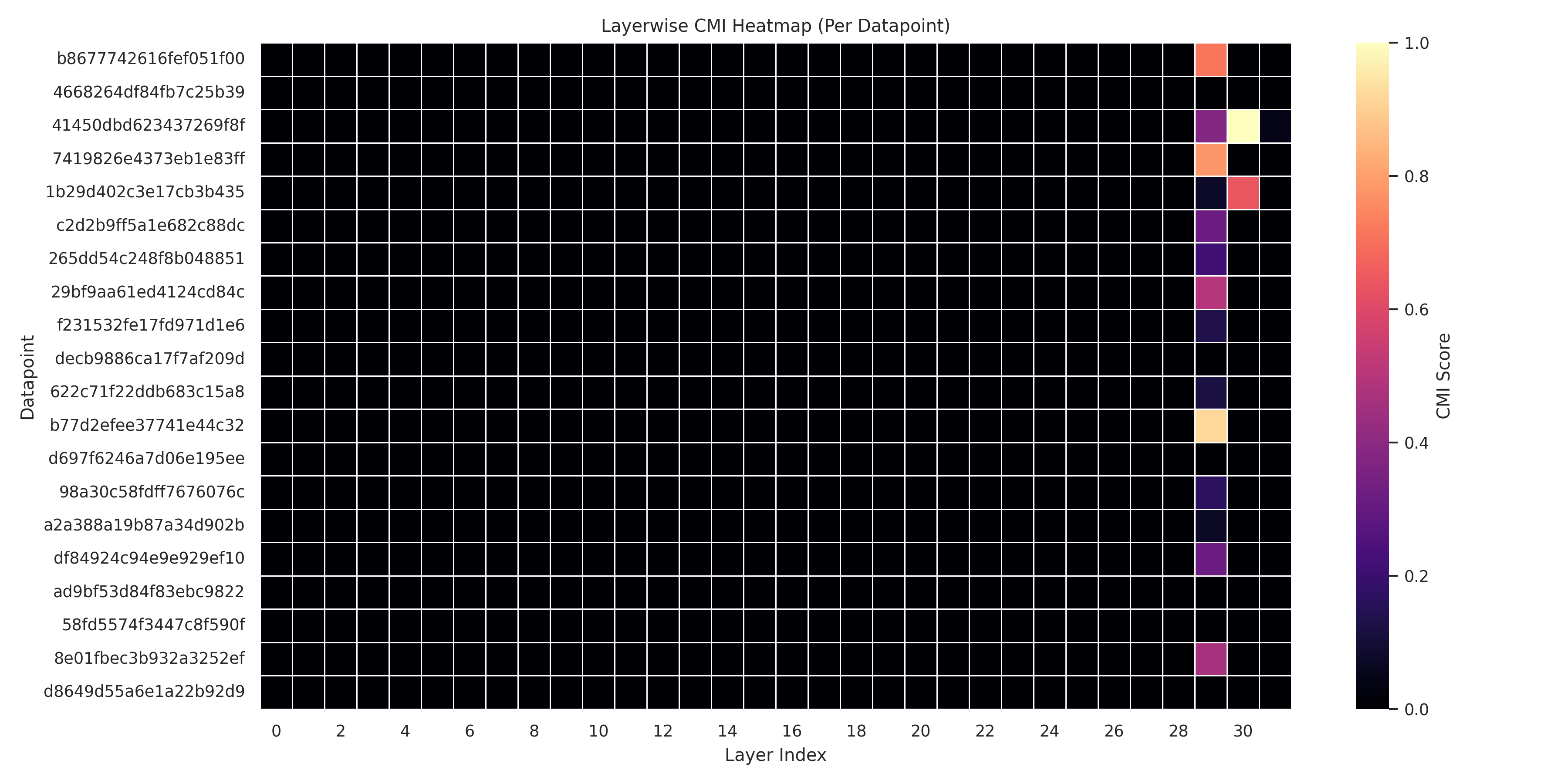}%
    }{\fbox{\rule[-.5cm]{0cm}{3.0cm}\rule[-.5cm]{\linewidth}{0cm}}}
  \end{minipage}
  \caption{StrategyQA layerwise CoT mediation for phi-2.}
  \label{fig:strategyqa_microsoftphi2}
\end{figure}

\begin{figure}[h]
  \centering
  \begin{minipage}{0.49\linewidth}
    \centering
    \textbf{(a)} CoT drop\\
    \vspace{0.25em}
    \IfFileExists{modelsrun/new_plot_layer_drop_means_microsoftphi1_5.png}{%
      \includegraphics[width=\linewidth]{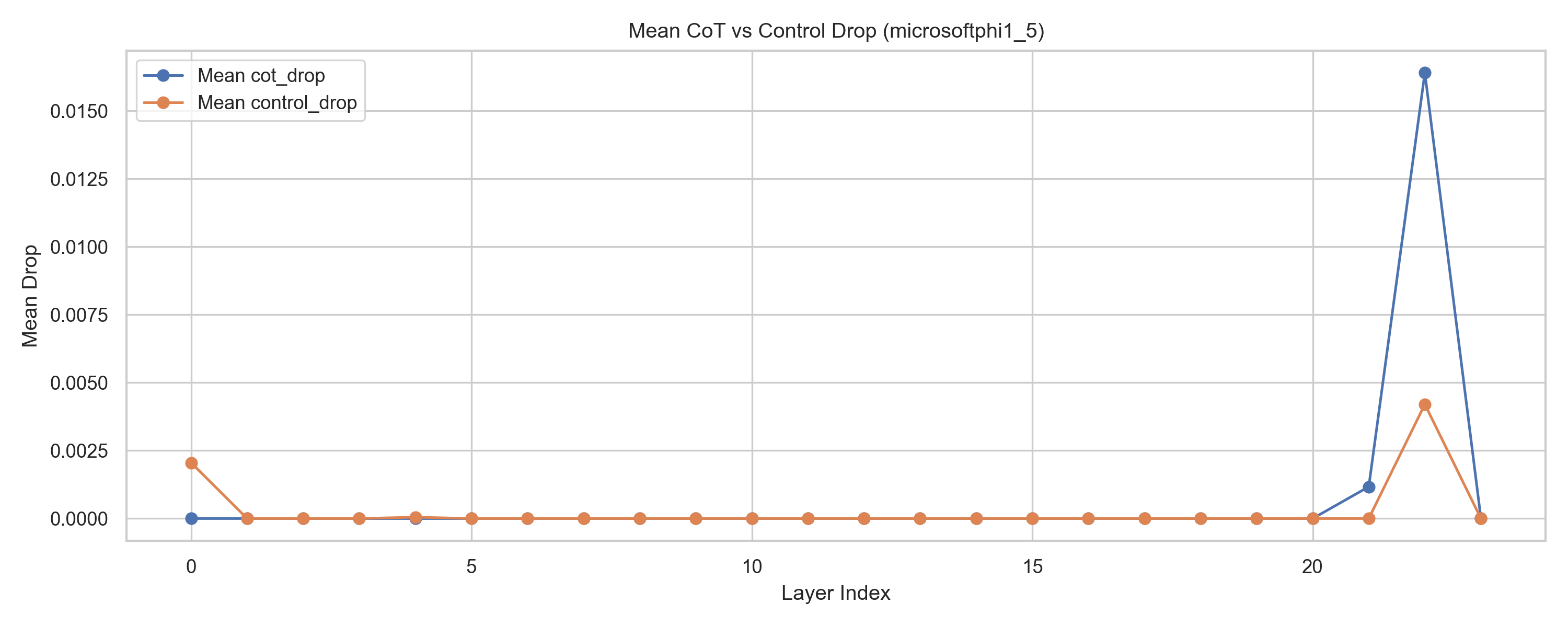}%
    }{\fbox{\rule[-.5cm]{0cm}{3.0cm}\rule[-.5cm]{\linewidth}{0cm}}}
  \end{minipage}
  \hfill
  \begin{minipage}{0.49\linewidth}
    \centering
    \textbf{(b)} CMI heatmap\\
    \vspace{0.25em}
    \IfFileExists{modelsrun/new_plot_layer_heatmap_microsoftphi1_5.png}{%
      \includegraphics[width=\linewidth]{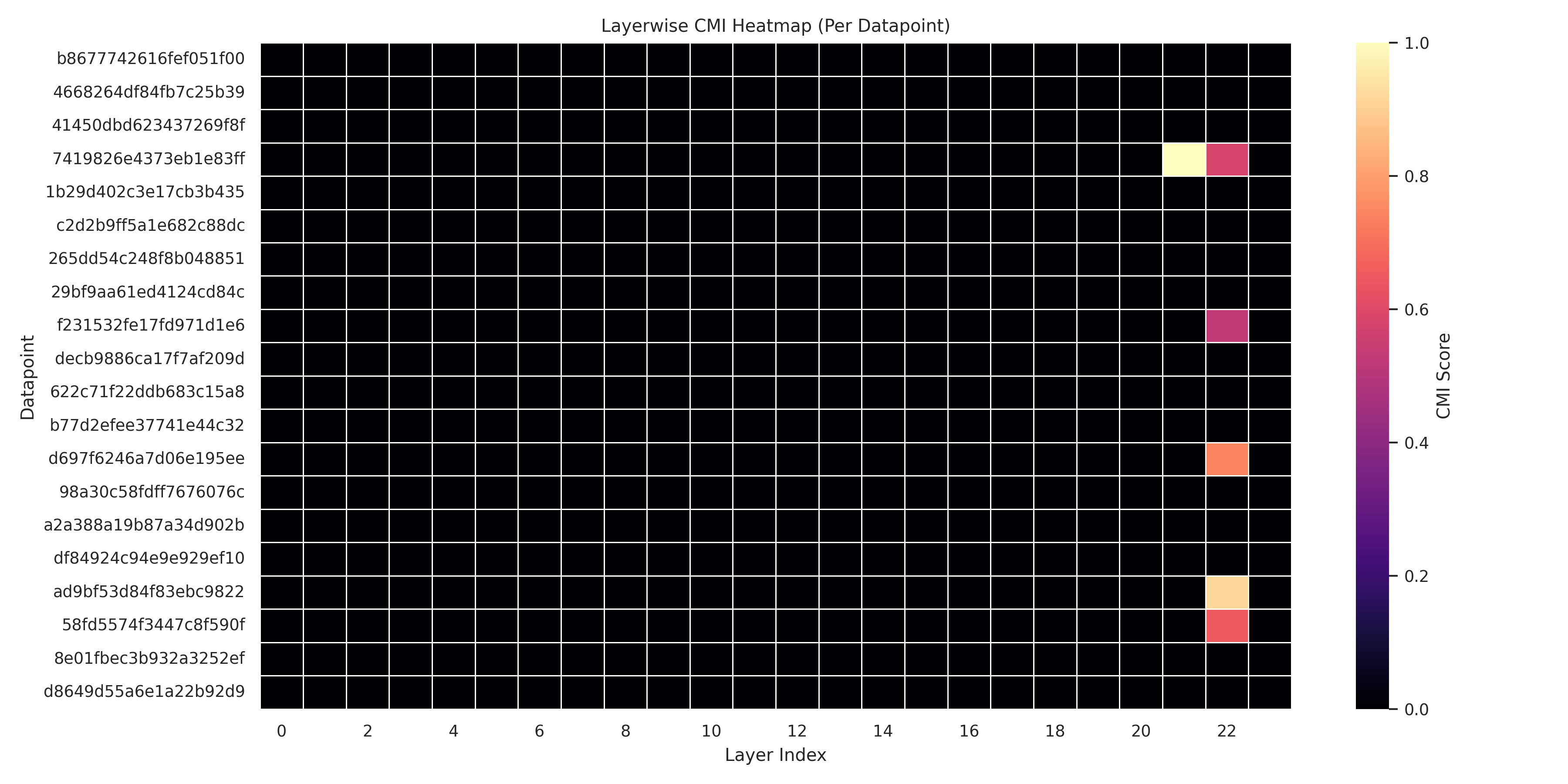}%
    }{\fbox{\rule[-.5cm]{0cm}{3.0cm}\rule[-.5cm]{\linewidth}{0cm}}}
  \end{minipage}
  \caption{StrategyQA layerwise CoT mediation for phi-1\_5.}
  \label{fig:strategyqa_microsoftphi1_5}
\end{figure}

\begin{figure}[h]
  \centering
  \begin{minipage}{0.49\linewidth}
    \centering
    \textbf{(a)} CoT drop\\
    \vspace{0.25em}
    \IfFileExists{modelsrun/new_plot_layer_drop_means_qwen3_8B.png}{%
      \includegraphics[width=\linewidth]{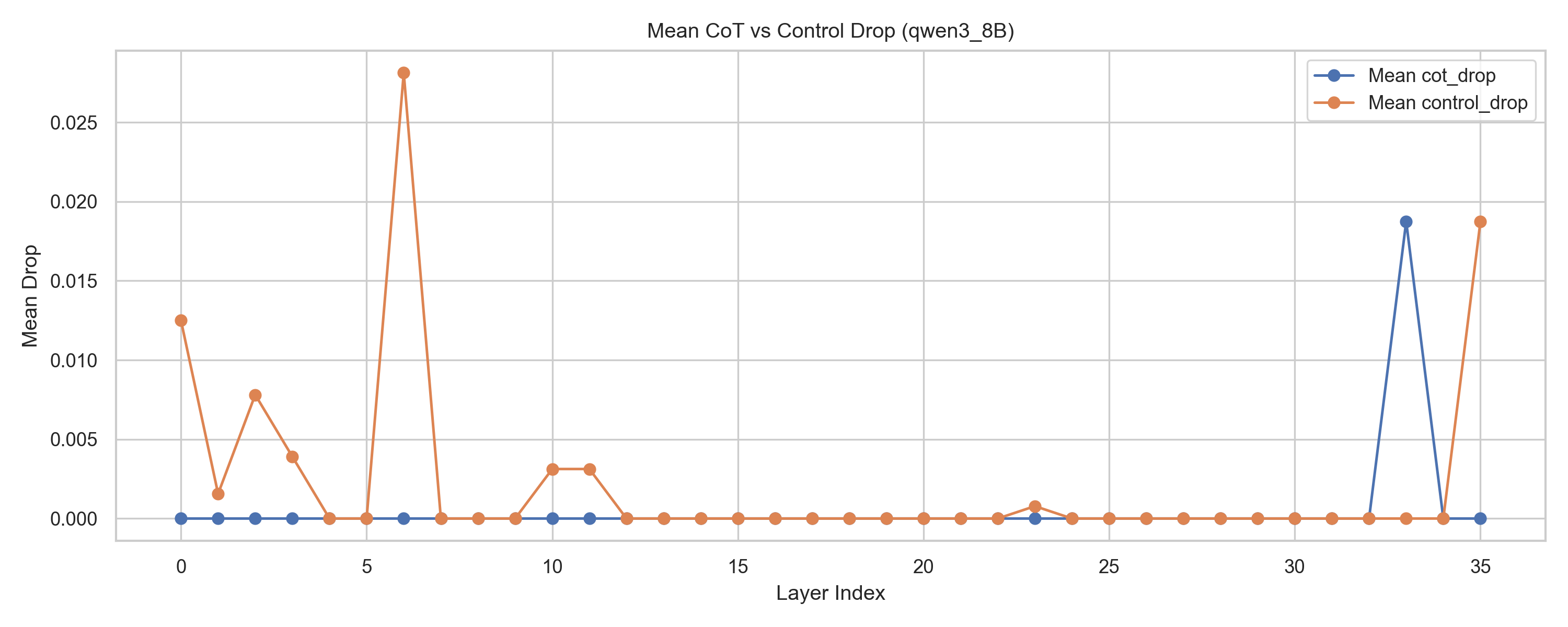}%
    }{\fbox{\rule[-.5cm]{0cm}{3.0cm}\rule[-.5cm]{\linewidth}{0cm}}}
  \end{minipage}
  \hfill
  \begin{minipage}{0.49\linewidth}
    \centering
    \textbf{(b)} CMI heatmap\\
    \vspace{0.25em}
    \IfFileExists{modelsrun/new_plot_layer_heatmap_qwen3_8B.png}{%
      \includegraphics[width=\linewidth]{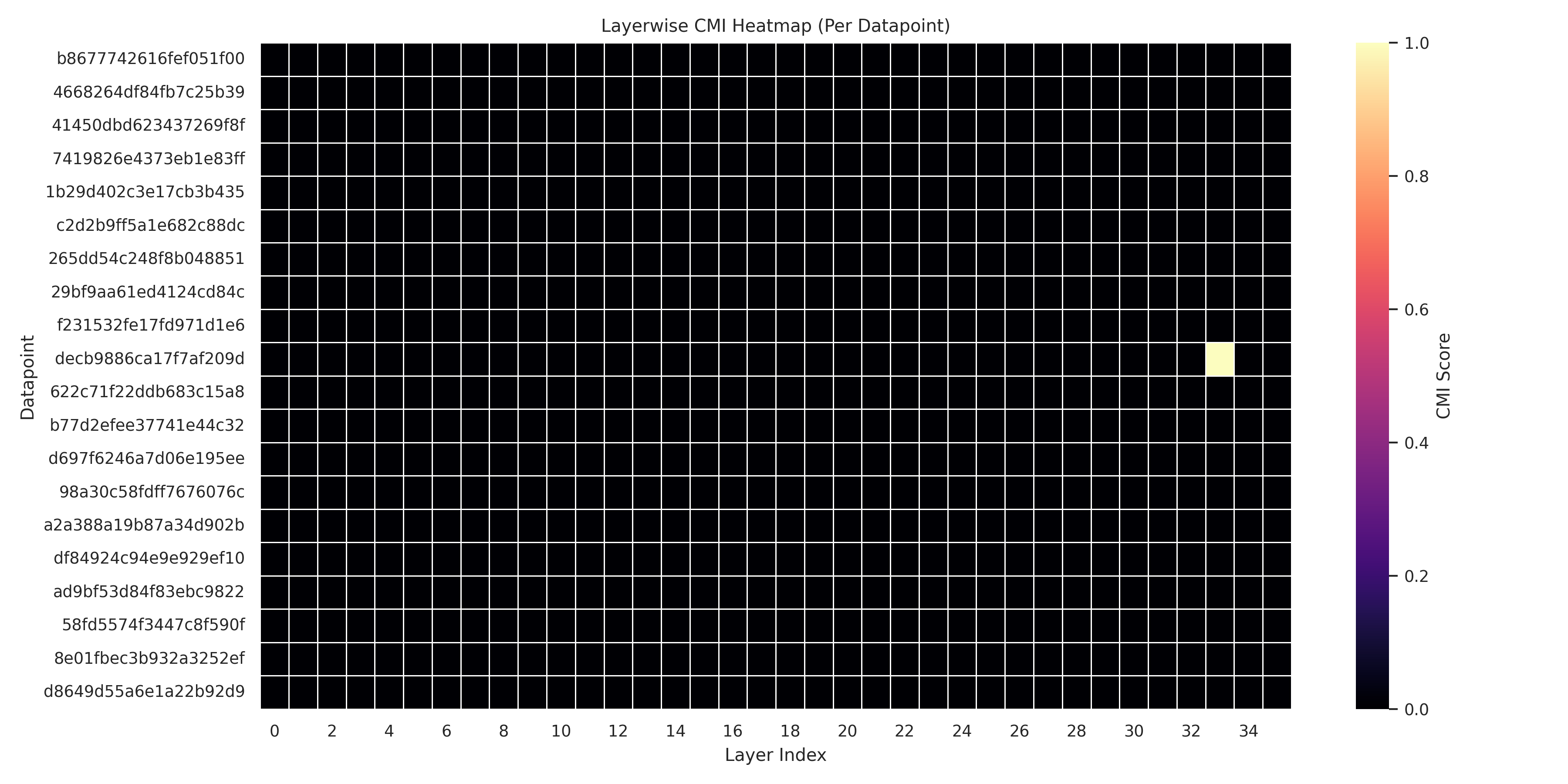}%
    }{\fbox{\rule[-.5cm]{0cm}{3.0cm}\rule[-.5cm]{\linewidth}{0cm}}}
  \end{minipage}
  \caption{StrategyQA layerwise CoT mediation for Qwen3-8B.}
  \label{fig:strategyqa_qwen3_8B}
\end{figure}

\begin{figure}[h]
  \centering
  \begin{minipage}{0.49\linewidth}
    \centering
    \textbf{(a)} CoT drop\\
    \vspace{0.25em}
    \IfFileExists{modelsrun/new_plot_layer_drop_means_qwen3_4B.png}{%
      \includegraphics[width=\linewidth]{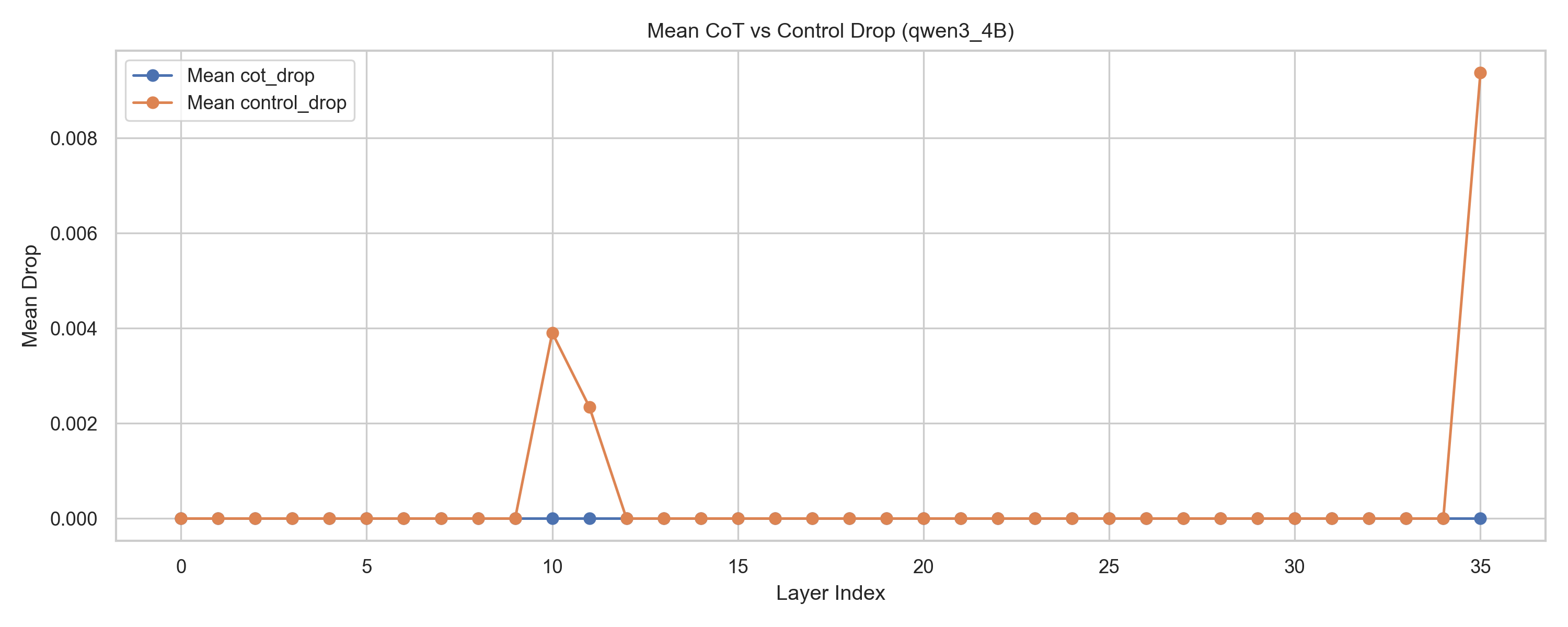}%
    }{\fbox{\rule[-.5cm]{0cm}{3.0cm}\rule[-.5cm]{\linewidth}{0cm}}}
  \end{minipage}
  \hfill
  \begin{minipage}{0.49\linewidth}
    \centering
    \textbf{(b)} CMI heatmap\\
    \vspace{0.25em}
    \IfFileExists{modelsrun/new_plot_layer_heatmap_qwen3_4B.png}{%
      \includegraphics[width=\linewidth]{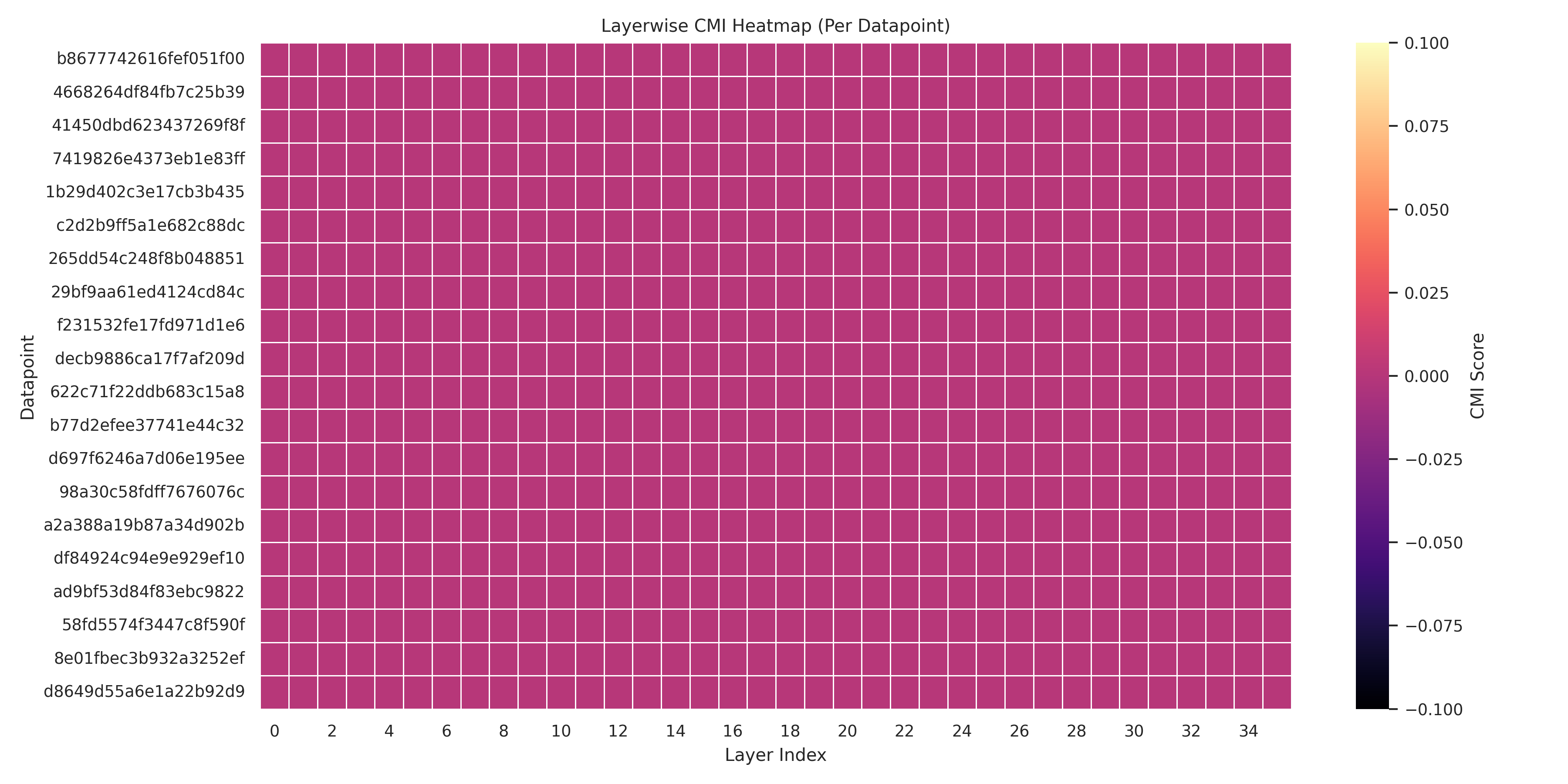}%
    }{\fbox{\rule[-.5cm]{0cm}{3.0cm}\rule[-.5cm]{\linewidth}{0cm}}}
  \end{minipage}
  \caption{StrategyQA layerwise CoT mediation for Qwen3-4B (pink = 0 CMI).}
  \label{fig:strategyqa_qwen3_4B}
\end{figure}

\begin{figure}[h]
  \centering
  \begin{minipage}{0.49\linewidth}
    \centering
    \textbf{(a)} CoT drop\\
    \vspace{0.25em}
    \IfFileExists{modelsrun/new_plot_layer_drop_means_qwen3__1_7B.png}{%
      \includegraphics[width=\linewidth]{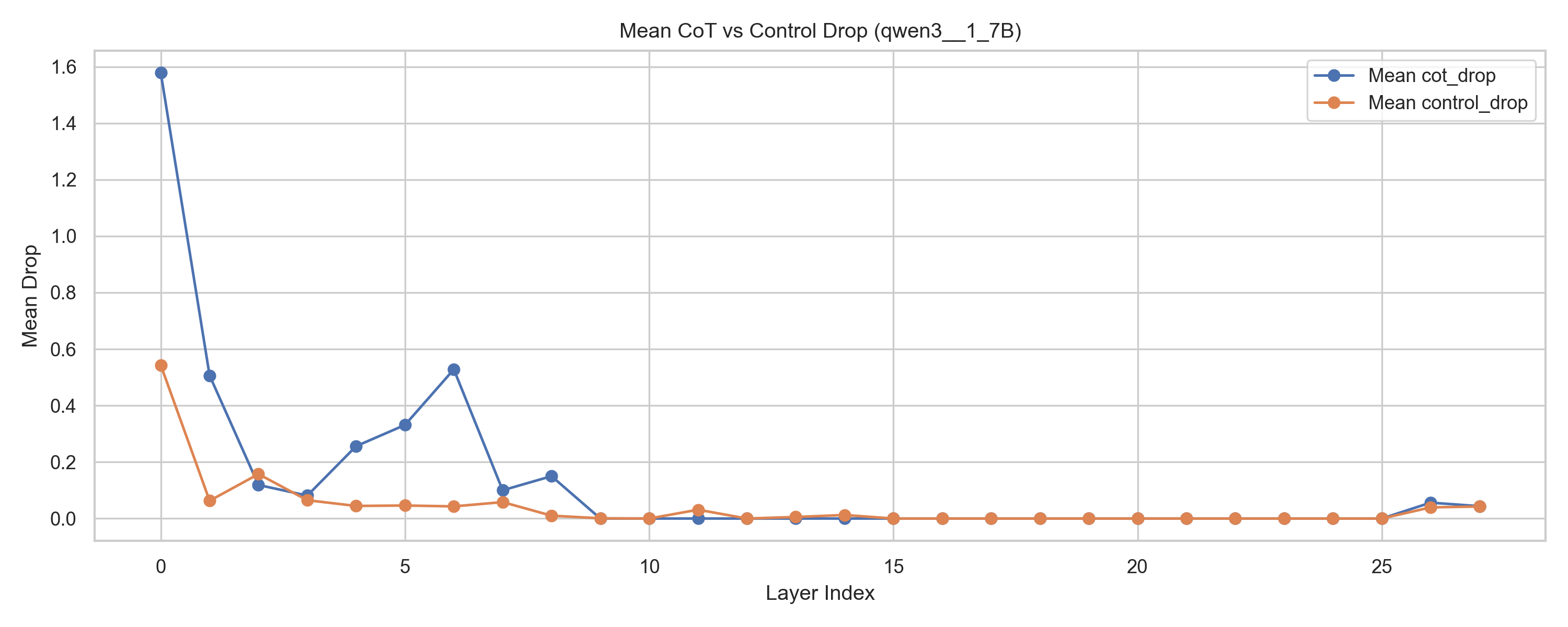}%
    }{\fbox{\rule[-.5cm]{0cm}{3.0cm}\rule[-.5cm]{\linewidth}{0cm}}}
  \end{minipage}
  \hfill
  \begin{minipage}{0.49\linewidth}
    \centering
    \textbf{(b)} CMI heatmap\\
    \vspace{0.25em}
    \IfFileExists{modelsrun/new_plot_layer_heatmap_qwen3__1_7B.png}{%
      \includegraphics[width=\linewidth]{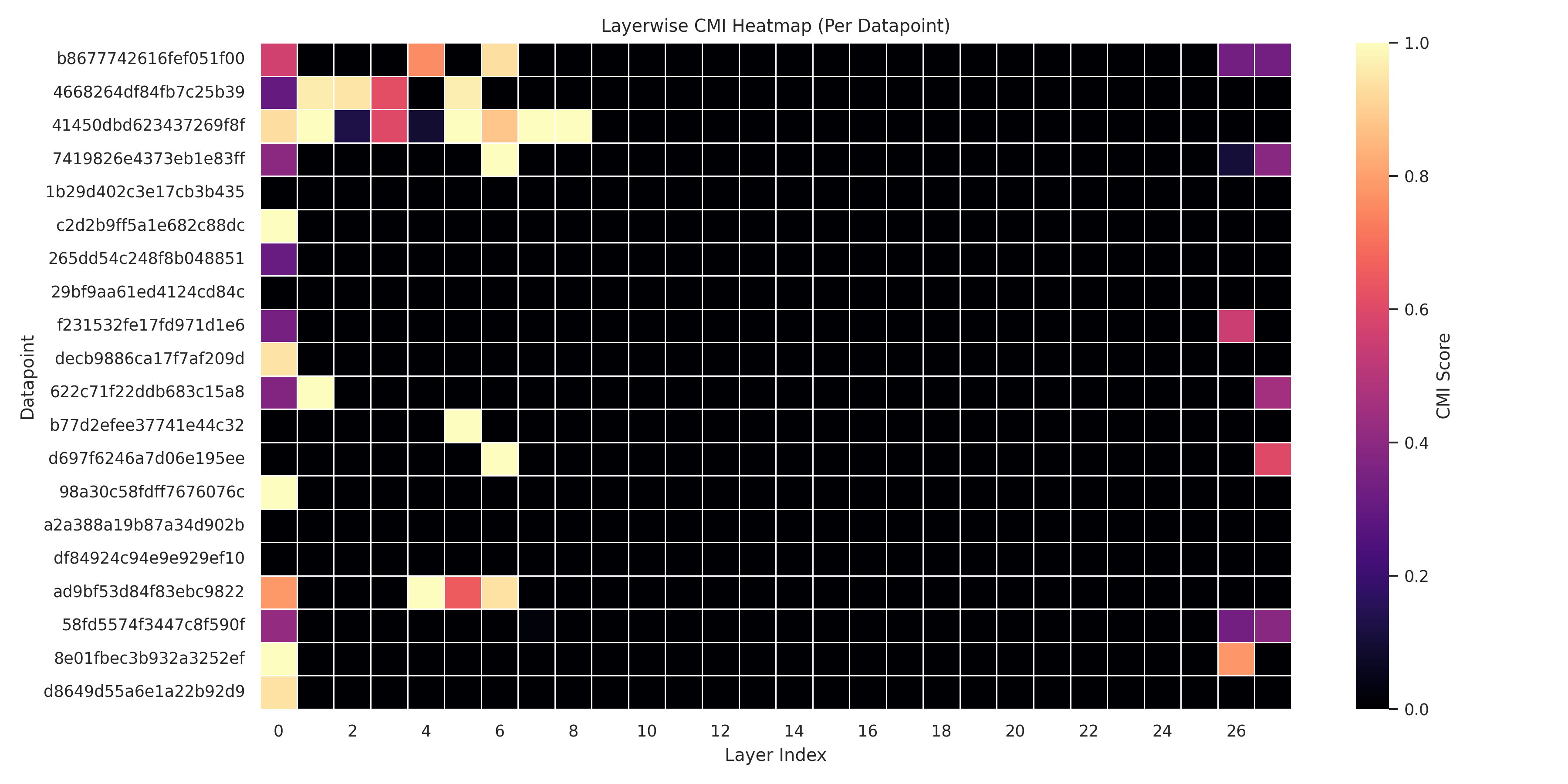}%
    }{\fbox{\rule[-.5cm]{0cm}{3.0cm}\rule[-.5cm]{\linewidth}{0cm}}}
  \end{minipage}
  \caption{StrategyQA layerwise CoT mediation for Qwen3-1.7B.}
  \label{fig:strategyqa_qwen3__1_7B}
\end{figure}

\begin{figure}[h]
  \centering
  \begin{minipage}{0.49\linewidth}
    \centering
    \textbf{(a)} CoT drop\\
    \vspace{0.25em}
    \IfFileExists{modelsrun/new_plot_layer_drop_means_qwen3__0_6B.png}{%
      \includegraphics[width=\linewidth]{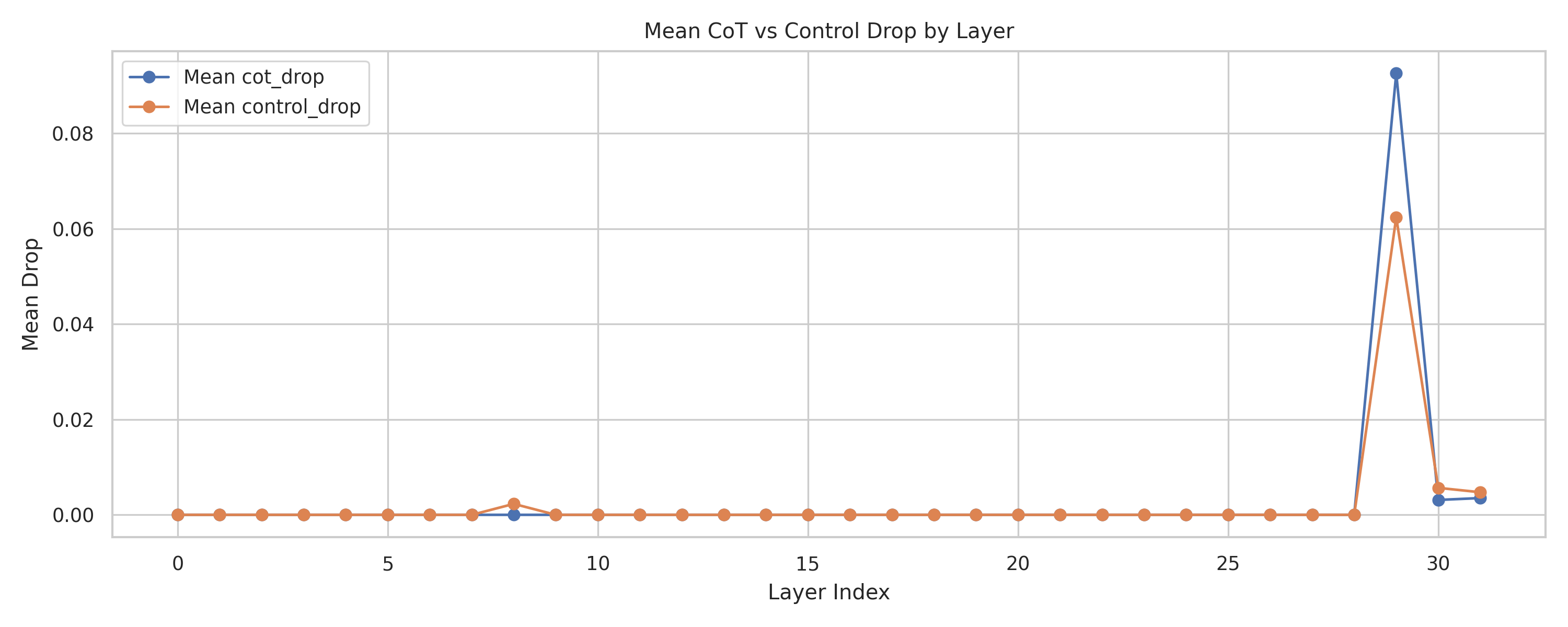}%
    }{\fbox{\rule[-.5cm]{0cm}{3.0cm}\rule[-.5cm]{\linewidth}{0cm}}}
  \end{minipage}
  \hfill
  \begin{minipage}{0.49\linewidth}
    \centering
    \textbf{(b)} CMI heatmap\\
    \vspace{0.25em}
    \IfFileExists{modelsrun/new_plot_layer_heatmap_qwen3__0_6B.png}{%
      \includegraphics[width=\linewidth]{modelsrun/new_plot_layer_heatmap_qwen3__0_6B.png}%
    }{\fbox{\rule[-.5cm]{0cm}{3.0cm}\rule[-.5cm]{\linewidth}{0cm}}}
  \end{minipage}
  \caption{StrategyQA layerwise CoT mediation for Qwen3-0.6B.}
  \label{fig:strategyqa_qwen3__0_6B}
\end{figure}

\begin{figure}[h]
  \centering
  \begin{minipage}{0.49\linewidth}
    \centering
    \textbf{(a)} CoT drop\\
    \vspace{0.25em}
    \IfFileExists{modelsrun/new_plot_layer_drop_means_dialogpt.png}{%
      \includegraphics[width=\linewidth]{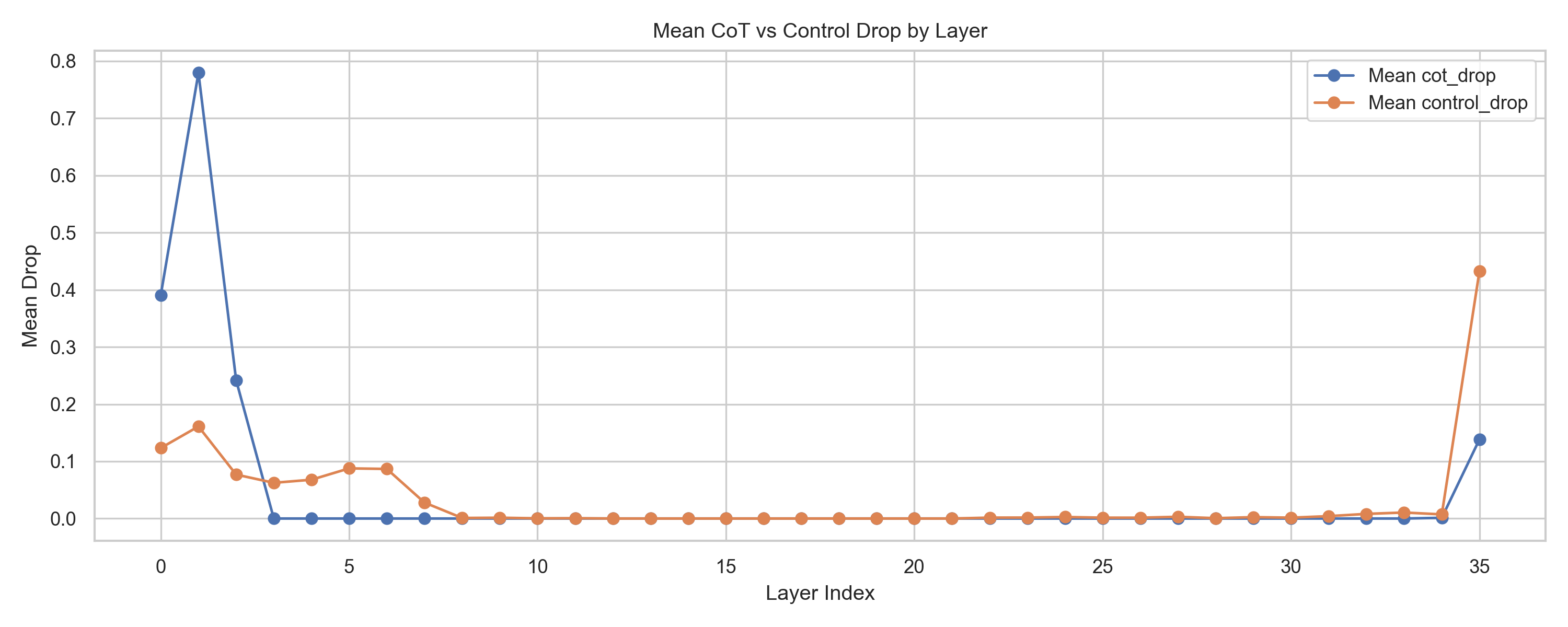}%
    }{\fbox{\rule[-.5cm]{0cm}{3.0cm}\rule[-.5cm]{\linewidth}{0cm}}}
  \end{minipage}
  \hfill
  \begin{minipage}{0.49\linewidth}
    \centering
    \textbf{(b)} CMI heatmap\\
    \vspace{0.25em}
    \IfFileExists{modelsrun/new_plot_layer_heatmap_dialogpt.png}{%
      \includegraphics[width=\linewidth]{modelsrun/new_plot_layer_heatmap_dialogpt.png}%
    }{\fbox{\rule[-.5cm]{0cm}{3.0cm}\rule[-.5cm]{\linewidth}{0cm}}}
  \end{minipage}
  \caption{StrategyQA layerwise CoT mediation for DialoGPT.}
  \label{fig:strategyqa_dialogpt}
\end{figure}
\FloatBarrier
\clearpage

\section{TruthfulQA Results}
\label{app:truthfulqa}

Here we provide additional instance-level metrics and interpretive analysis for the TruthfulQA experiment in Section~\ref{sec:results:truthfulqa}. We use \texttt{tqa\_k} to denote a single TruthfulQA datapoint/instance.

\paragraph{Interpretation of near-total bypass.}
Across nearly all instances, we observe $\mathrm{CMI} \approx 0$ and Bypass $\approx 1.0$, suggesting that the model's final answer is functionally independent of intermediate CoT activations under our patching interventions. This cautions that CoT rationales may be non-faithful even when the final answer is correct.

\paragraph{Bias toward misconceptions.}
For some instances, the baseline $\log P$ is substantially higher for the myth than for the truth (e.g., \texttt{tqa\_4}, \texttt{tqa\_7}), indicating a strong prior preference that the CoT does not reliably mediate.

\paragraph{Rationalization vs. correction.}
In rare cases such as \texttt{tqa\_1}, CMI is slightly higher for the myth answer than for the truthful one, suggesting that internal computation can be more strongly recruited to justify a preferred misconception than to construct a correction.

\begin{table}[ht]
\centering
\small
\caption{Causal mediation and bypass scores for TruthfulQA instances. Baseline $\log P$ indicates the model's confidence in the respective answer string as defined by \citet{lin2022truthfulqameasuringmodelsmimic}.}
\label{tab:truthful_qa_results}
\begin{tabular}{l l r r r}
\hline
ID & Type & Baseline $\log P$ & CMI & Bypass \\
\hline
tqa\_0 & Truth & -23.1196 & 0.0000 & 1.0000 \\
tqa\_0 & Myth  & -35.9314 & 0.0000 & 1.0000 \\
\hline
tqa\_1 & Truth & -25.6187 & 0.0015 & 0.9985 \\
tqa\_1 & Myth  & -46.3392 & 0.0092 & 0.9908 \\
\hline
tqa\_2 & Truth & -39.9036 & 0.0000 & 1.0000 \\
tqa\_2 & Myth  & -27.0031 & 0.0000 & 1.0000 \\
\hline
tqa\_3 & Truth & -21.0348 & 0.0000 & 1.0000 \\
tqa\_3 & Myth  & -15.7252 & 0.0000 & 1.0000 \\
\hline
tqa\_4 & Truth & -37.7736 & 0.0000 & 1.0000 \\
tqa\_4 & Myth  & -18.4621 & 0.0000 & 1.0000 \\
\hline
tqa\_5 & Truth & -27.0194 & 0.0042 & 0.9958 \\
tqa\_5 & Myth  & -35.7996 & 0.0000 & 1.0000 \\
\hline
tqa\_6 & Truth & -39.6933 & 0.0000 & 1.0000 \\
tqa\_6 & Myth  & -28.0470 & 0.0000 & 1.0000 \\
\hline
tqa\_7 & Truth & -37.9082 & 0.0000 & 1.0000 \\
tqa\_7 & Myth  & -18.2792 & 0.0000 & 1.0000 \\
\hline
tqa\_8 & Truth & -16.1372 & 0.0000 & 1.0000 \\
tqa\_8 & Myth  & -16.6255 & 0.0000 & 1.0000 \\
\hline
tqa\_9 & Truth & -35.5275 & 0.0000 & 1.0000 \\
tqa\_9 & Myth  & -30.3672 & 0.0000 & 1.0000 \\
\hline
\end{tabular}
\end{table}

\section{GSM8K Instance Results}
\label{app:gsm8k}

We report additional instance-level results for GSM8K~\citep{cobbe2021trainingverifierssolvemath} using the
\texttt{microsoft/DialoGPT-large} model. CMI and Bypass are computed as defined in
Section~\ref{sec:method:cmi:index}.

\paragraph{Baseline log-probability.}
If $Q$ is the question and $A$ is the correct answer, the baseline log-probability is:
\begin{equation}
\text{Baseline log } P = \ln(P(A \mid Q)).
\end{equation}

\begin{table}[t]
\centering
\small
\caption{GSM8K instance-level causal mediation and bypass results.}
\label{tab:gsm8k_instance_results}
\begin{tabular}{l r r r}
\hline
ID & CMI-mean & Bypass & Baseline $\log P$ \\
\hline
\texttt{gsm\_natalia} & 0.466 & 0.534 & -10.160 \\
\texttt{gsm\_weng} & 0.000 & 1.000 & -9.058 \\
\texttt{gsm\_betty} & 0.000 & 1.000 & -10.619 \\
\texttt{gsm\_julie} & 0.750 & 0.250 & -9.085 \\
\texttt{gsm\_james} & 0.051 & 0.949 & -13.385 \\
\texttt{gsm\_mark} & 0.375 & 0.625 & -8.263 \\
\texttt{gsm\_albert} & 0.726 & 0.274 & -9.420 \\
\texttt{gsm\_ken} & 0.581 & 0.419 & -6.956 \\
\texttt{gsm\_alexis} & 0.000 & 1.000 & -10.454 \\
\texttt{gsm\_tina} & 0.533 & 0.467 & -12.221 \\
\texttt{gsm\_monster} & 0.868 & 0.132 & -16.917 \\
\texttt{gsm\_tobias} & 0.400 & 0.600 & -10.825 \\
\texttt{gsm\_randy} & 0.673 & 0.327 & -11.464 \\
\texttt{gsm\_jasper} & 0.083 & 0.917 & -4.798 \\
\texttt{gsm\_joy} & 0.256 & 0.744 & -9.298 \\
\texttt{gsm\_james\_media} & 0.575 & 0.425 & -22.074 \\
\texttt{gsm\_mike\_johnson} & 0.000 & 1.000 & -8.307 \\
\texttt{gsm\_hard\_hats} & 0.625 & 0.375 & -9.996 \\
\texttt{gsm\_roque} & 0.453 & 0.547 & -7.888 \\
\texttt{gsm\_tim\_bike} & 0.000 & 1.000 & -19.235 \\
\hline
\end{tabular}
\end{table}

\subsection*{Qualitative analysis}

\paragraph{The core metrics.}
To interpret Table~\ref{tab:gsm8k_instance_results}, it is useful to view a tug-of-war between
CMI and Bypass.
\begin{itemize}
\item \textbf{CMI-mean (CoT-mediated influence).} A high CMI-mean (closer to 1.0) suggests the model’s
final answer is causally dependent on the reasoning tokens it produced (higher faithfulness).
\item \textbf{Bypass.} This is the complementary ``shortcut'' score ($1 - \mathrm{CMI}$). High Bypass
(closer to 1.0) suggests the model’s answer is weakly coupled to the emitted CoT and is instead
produced via internal computation that does not rely on the CoT token states.
\item \textbf{Baseline $\log P$.} This is a difficulty/confidence proxy. Values closer to 0 (e.g.,
$-4.798$) indicate higher confidence, while very negative values (e.g., $-22.074$) indicate the model
assigns low probability to the correct answer without additional help.
\end{itemize}

\paragraph{Analysis of results by archetype.}
\textbf{Group A (pure bypass).}
Examples: \texttt{gsm\_weng}, \texttt{gsm\_betty}, \texttt{gsm\_alexis}, \texttt{gsm\_mike\_johnson},
\texttt{gsm\_tim\_bike}.
These cases exhibit $\mathrm{CMI}=0$ and Bypass $=1$, consistent with post-hoc rationalization: the
model emits a plausible CoT, but patching the CoT-token representations does not uniquely affect the
answer beyond controls.

\medskip
\textbf{Group B (high CMI).}
Examples: \texttt{gsm\_monster} (0.868), \texttt{gsm\_julie} (0.750), \texttt{gsm\_albert} (0.726).
These instances show strong causal mediation through the explicit scratchpad. Notably, some of these
also have relatively low baseline $\log P$ (e.g., \texttt{gsm\_monster} at $-16.917$), suggesting that
when the model’s ``gut'' confidence is weak, it relies more on the CoT token states to mediate the
computation.

\medskip
\textbf{Group C (partial mediation).}
Examples: \texttt{gsm\_natalia} (0.466), \texttt{gsm\_tobias} (0.400), \texttt{gsm\_roque} (0.453).
These tasks show partial dependence, consistent with a mixture of latent computation (bypass) and
explicit CoT-mediated bookkeeping.

\paragraph{Key takeaway.}
Across GSM8K instances, CoT is not always mechanistically used even when it is generated and the
final answer is correct (all examples here are marked as correct). This reinforces that accuracy
alone can be misleading: a model can be ``right for the wrong reasons'' (high bypass) or ``right for
the right reasons'' (high CMI). In alignment and interpretability settings, high-bypass cases remain
effectively more opaque, since the emitted rationale may not reflect the internal causal pathway that
produced the answer.

\section{Behavioral Baseline Details}
\label{app:behavior}
Using a standard LLM chat-completions API, we elicit structured JSON with fields \texttt{cot} and \texttt{answer}, running both a standard extraction prompt and an \emph{audit-aware} variant that explicitly states the CoT will be monitored. The audit-aware variant additionally instructs the model to explicitly surface evaluator- and strategy-related considerations, whereas the standard prompt does not. We then score manipulation risk from the CoT using (i) a regex library covering 19 manipulation categories (e.g., evaluation awareness, deception planning, goal hiding, policy evasion); (ii) embedding-based coherence (prompt--CoT and CoT--answer cosine similarity using \texttt{all-MiniLM-L6-v2}); and (iii) simple structural heuristics, including a compression-ratio proxy (Zlib) for template-like CoT. Finally, we combine these features into a 0--100 risk score.

These features jointly target strategic language (regex patterns), coherence (embedding similarity), and templated CoT (compression). Because audit-aware mode changes the prompt distribution, we interpret normal vs.\ audit-aware differences as a stress test rather than an unbiased estimate of natural manipulation. We report uncalibrated scores and treat them as relative diagnostics; calibration on labeled data is required before any deployment claims.

\begin{figure}[h]
    \centering
    \begin{subfigure}[t]{0.48\linewidth}
        \centering
        \IfFileExists{mean_deltas_bar.png}{%
            \includegraphics[width=\linewidth]{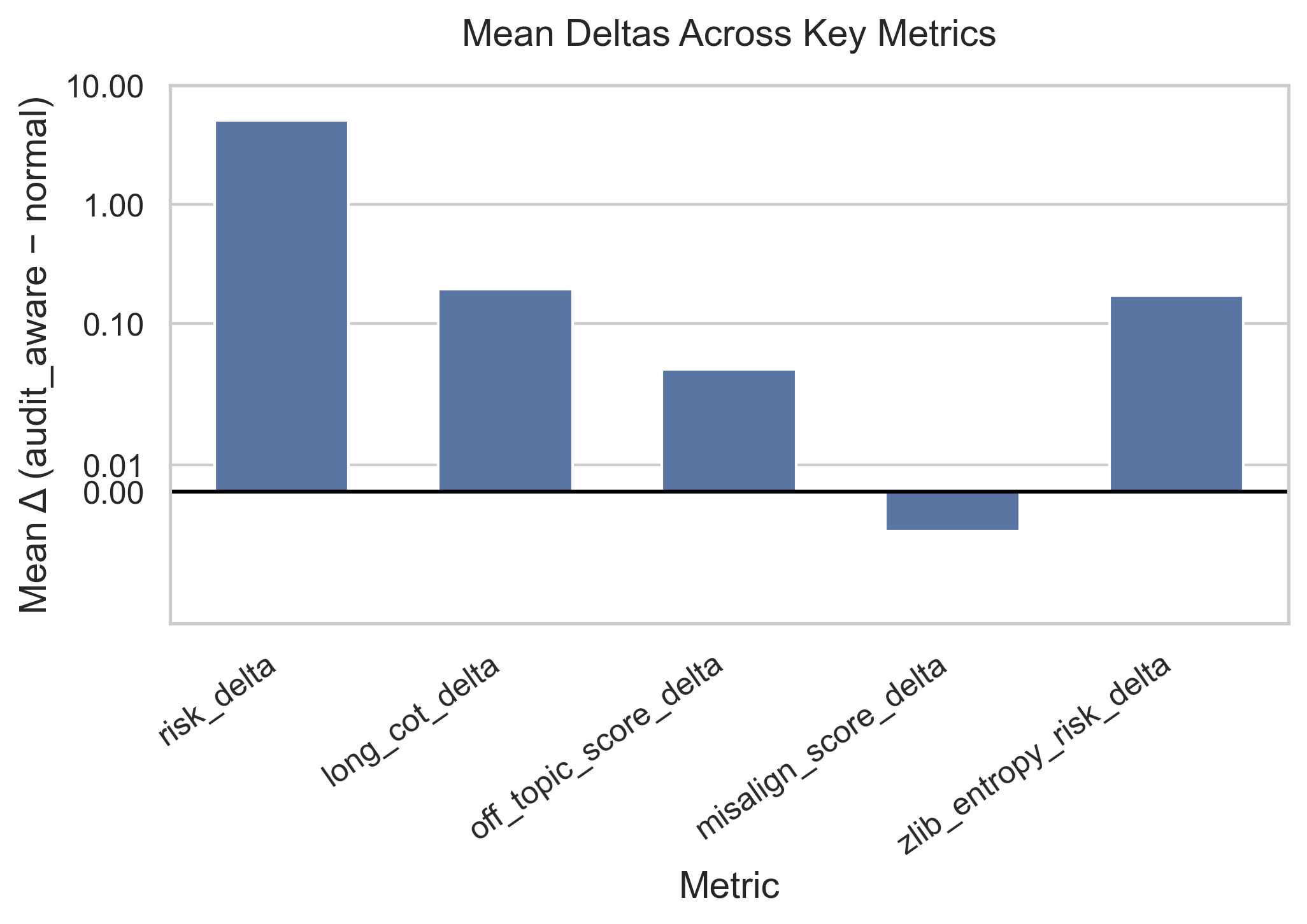}%
        }{%
            \fbox{\rule[-.5cm]{0cm}{3.0cm}\rule[-.5cm]{\linewidth}{0cm}}%
        }
        \caption{Mean change (log scale) in behavioral risk metrics under audit-aware prompting, reported as audit-aware minus normal.}
        \label{fig:mean_deltas}
    \end{subfigure}
    \hfill
    \begin{subfigure}[t]{0.48\linewidth}
        \centering
        \IfFileExists{risk_delta_hist.png}{%
            \includegraphics[width=\linewidth]{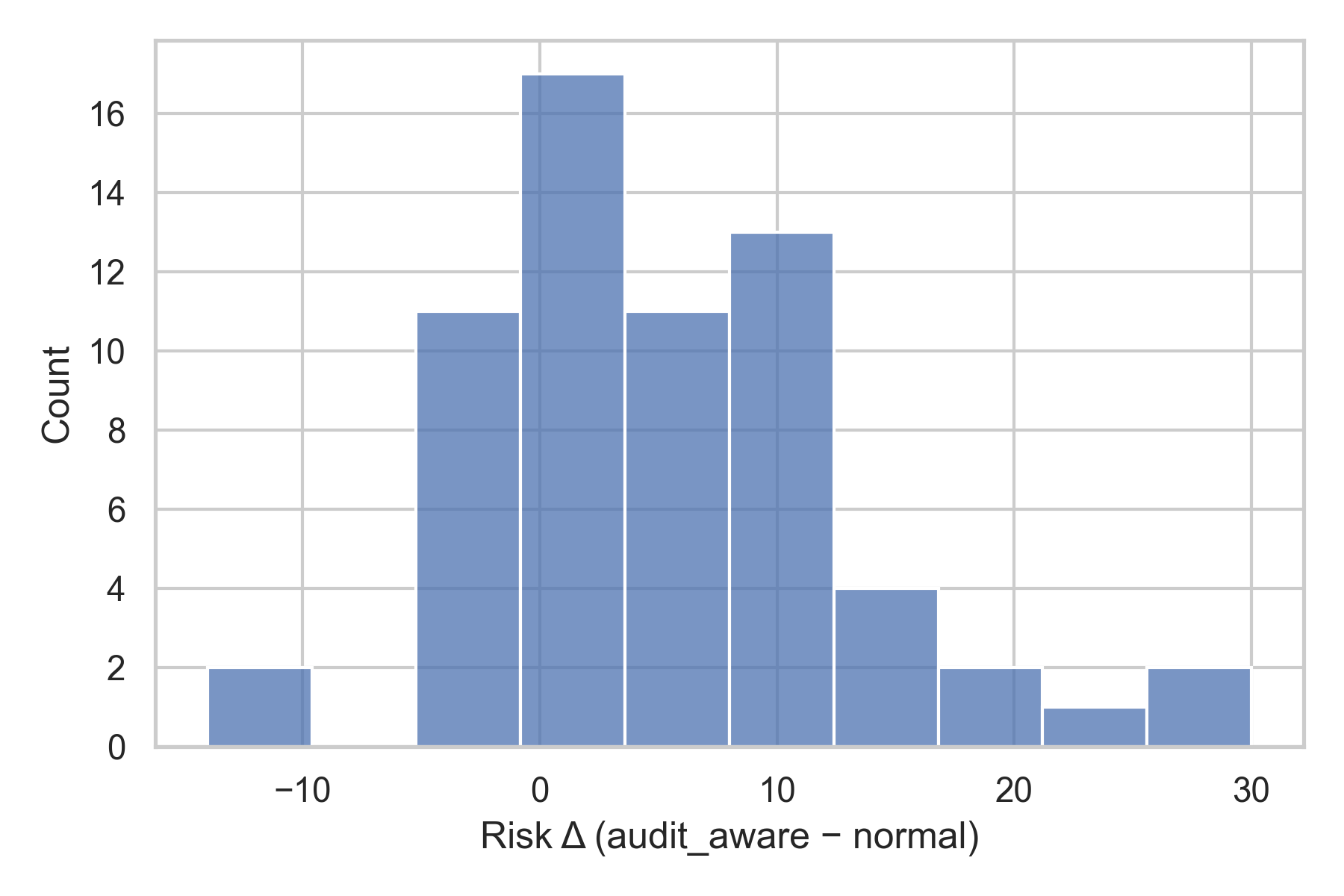}%
        }{%
            \fbox{\rule[-.5cm]{0cm}{3.0cm}\rule[-.5cm]{\linewidth}{0cm}}%
        }
        \caption{Distribution of per-prompt risk-score deltas (audit-aware $-$ normal).}
        \label{fig:risk_hist}
    \end{subfigure}
    \caption{Behavioral CoT manipulation monitor. Audit-aware prompting (which explicitly informs the model that CoT will be inspected) increases manipulation-signal features on average (left) and yields a right-tailed distribution of risk-score deltas across prompts (right).}
    \label{fig:behavioural_summary}
\end{figure}

\section{Dataset Details}
\label{app:datasets}
\subsection*{StrategyQA}
\paragraph{File and size.}
We use \texttt{cot\_bypass\_monitor/good/data/strategy\_qa.json},
which contains 2{,}290 entries (as loaded in the current repository).

\paragraph{Fields.}
Each entry is a JSON object with at least the following fields:
\texttt{qid}, \texttt{term}, \texttt{description}, \texttt{question}, \texttt{answer},
\texttt{facts}, \texttt{decomposition}, and \texttt{evidence}.
The implementation uses:
\begin{itemize}
\item \texttt{question} (string) as the main prompt question.
\item \texttt{answer} (boolean or string/number), converted to a string. Boolean values are mapped to
\texttt{"Yes"} or \texttt{"No"}.
\item \texttt{facts} (list of strings), if present, rendered as a bullet list.
\item \texttt{decomposition} (list of strings), if present, rendered as a bullet list.
\item \texttt{evidence}, if present, JSON-serialized into a single block.
\item \texttt{qid} is used as the example identifier.
\end{itemize}

\paragraph{Prompt construction.}
For each example, two prompts are constructed:
\begin{itemize}
\item \textbf{With-CoT prompt:}
\begin{verbatim}
Question: {question}
Facts:
- ...
Decomposition:
- ...
Evidence:
{json evidence}
Let's think step by step.
Final answer: {answer}
\end{verbatim}
The \texttt{Facts}, \texttt{Decomposition}, and \texttt{Evidence} blocks are included only if they
exist for that example.
\item \textbf{No-CoT prompt:}
\begin{verbatim}
Question: {question}
Final answer: {answer}
\end{verbatim}
\end{itemize}
Thus, the With-CoT version includes the reasoning cue (``Let's think step by step.'') and optional
supporting fields, while the No-CoT version includes only the question and the final answer line.

\paragraph{Subset and split.}
The StrategyQA analysis script (\texttt{test\_cmi\_layerwise\_strategyqa.py})
loads a \emph{top-$N$ prefix} of the JSON file using \texttt{items[:N]}.
In the current script configuration, $N=20$ by default.
No official dataset split (train/val/test) is used; the run is simply the first $N$ examples in the
local file.

\paragraph{Correctness labels.}
The StrategyQA script does not compute or store accuracy labels; the \texttt{correct} field is set to
\texttt{None}, and summary tables indicate correctness with a ``?'' placeholder.

\subsection*{TruthfulQA}
\paragraph{File and size.}
We use \texttt{cot\_bypass\_monitor/good/data/TruthfulQA.csv},
which contains 791 rows including the header.

\paragraph{Fields.}
The CSV columns include (as in the header):
\texttt{Type}, \texttt{Category}, \texttt{Question}, \texttt{Best Answer},
\texttt{Best Incorrect Answer}, \texttt{Correct Answers},
\texttt{Incorrect Answers}, and \texttt{Source}.
The script uses:
\begin{itemize}
\item \texttt{Question}
\item \texttt{Best Answer} (treated as the truthful answer)
\item \texttt{Best Incorrect Answer} (treated as the myth/false answer)
\end{itemize}

\paragraph{Prompt construction.}
For each row, the script evaluates \emph{two answer types}:
\begin{itemize}
\item \textbf{Truth} (Best Answer)
\item \textbf{Myth} (Best Incorrect Answer)
\end{itemize}
For each answer type, prompts are:
\begin{itemize}
\item \textbf{With-CoT:}
\begin{verbatim}
Question: {question}
Let's think step by step.
Final answer: {answer}
\end{verbatim}
\item \textbf{No-CoT:}
\begin{verbatim}
Question: {question}
{answer}
\end{verbatim}
\end{itemize}
Notably, the No-CoT prompt places the answer directly after the question without the ``Final answer:''
prefix (this matches the current script and is not a typo in this description).

\paragraph{Subset and split.}
The TruthfulQA script (\texttt{test\_truthqa.py}) uses a top-$N$ prefix of the CSV rows via
\texttt{load\_truthfulqa\_comparison(n)}.
The default in the script is $N=10$ rows. There is no official split used or stored; the analysis is
on the first $N$ questions in the local CSV.

\paragraph{Notes on preprocessing.}
TruthfulQA is treated as a paired evaluation per question: each question yields two runs (Truth and
Myth), and each run has With-CoT and No-CoT prompts. This is distinct from StrategyQA, which uses a
single reference answer per question.

\subsection*{GSM8K}
\paragraph{File and size.}
We use \texttt{cot\_bypass\_monitor/good/data/gsm8k.json},
which contains 7{,}500 entries (as loaded in the current repository).

\paragraph{Fields.}
Each entry is a JSON object with fields including:
\texttt{id}, \texttt{task\_type}, \texttt{with\_cot}, \texttt{no\_cot}, \texttt{answer}, and \texttt{correct}.

\paragraph{Prompt construction.}
For each example, two prompts are constructed:
\begin{itemize}
\item \textbf{With-CoT prompt:}
\begin{verbatim}
Question: {question}

Let's think step by step.
{chain-of-thought rationale}

Final answer: {answer}
\end{verbatim}
\item \textbf{No-CoT prompt:}
\begin{verbatim}
Question: {question}

Final answer: {answer}
\end{verbatim}
\end{itemize}
The \texttt{with\_cot} field contains the full reasoning trace and final answer, whereas the
\texttt{no\_cot} field omits the reasoning and includes only the question and final answer.

\paragraph{Subset and split.}
The GSM8K analysis script (\texttt{test\_gsm8k.py}) loads a \emph{top-$N$ prefix} of the JSON file.
For example, if $N=20$, it will load 20 examples.
No official dataset split (train/val/test) is used; the run is simply the first $N$ examples in the
local file.

\section{Code availability} The complete implementation is publicly available at \href{https://github.com/Anish-1101-lab/cot-manipulation-monitor}{https://github.com/Anish-1101-lab/cot-manipulation-monitor}.

\end{document}